\documentclass[runningheads]{llncs}

\usepackage[mobile]{eccv}

\usepackage{eccvabbrv}

\usepackage{graphicx}
\usepackage{booktabs}
\usepackage{nicefrac}
\usepackage{enumitem}
\usepackage{colortbl}
\usepackage{amssymb}
\usepackage{multirow}
\usepackage{makecell}

\usepackage{siunitx}  %
\DeclareSIUnit{\million}{\text{M}}
\ifdefined\pdfcompresslevel
  \usepackage[accsupp]{axessibility}  %
\else
  \PackageWarning{defs}{Skipping axessibility because \string\pdfcompresslevel is unavailable}
\fi

\usepackage[breaklinks,colorlinks,citecolor=eccvblue,urlcolor=eccvblue]{hyperref}

\usepackage[capitalize]{cleveref}

\newcommand{\ours}{VidMap}

\definecolor{bearingcolor}{RGB}{0,95,180}     %
\definecolor{depthcolor}{RGB}{0,140,70}       %
\definecolor{scalecolor}{RGB}{215,95,0}       %

\newcommand{\colbearing}[1]{{\color{bearingcolor}#1}}
\newcommand{\coldepth}[1]{{\color{depthcolor}#1}}
\newcommand{\colscale}[1]{{\color{scalecolor}#1}}

\newcommand{\0}{\phantom{0}}

\def \customparskip {4pt} %
\renewcommand{\paragraph}[1]{\vspace{\customparskip}\noindent\textbf{#1}}

  \definecolor{tabfirst}{HTML}{7ABDAF}
  \definecolor{tabsecond}{HTML}{BBDDB4}
  \definecolor{tabthird}{HTML}{EAF5A8}
  \definecolor{tabinvalid}{HTML}{E6E6E6}

\newcommand{\cfirst}{\cellcolor{tabfirst}\bfseries}
\newcommand{\csecond}{\cellcolor{tabsecond}}
\newcommand{\cthird}{\cellcolor{tabthird}}
\newcommand{\cinvalid}[1]{\cellcolor{tabinvalid}\strut\textcolor{red!70!black}{#1}}

\begin{document}

\title{VidMap: Exploiting Temporal Structure\\for Video-Based Structure-from-Motion}

\titlerunning{VidMap}

\author{%
Zador Pataki$^{1}$
\hspace{.15in}
Paul-Edouard Sarlin$^{2}$
\hspace{.15in}
Marc Pollefeys$^{1,3}$
\\
\vspace{.1in}
$^{1}$ETH Zurich
\hspace{.2in}%
$^{2}$Google
\hspace{.2in}%
$^{3}$Microsoft Spatial AI Lab
}

\authorrunning{Pataki \etal}

\institute{}

\maketitle

\begin{abstract}
Accurately recovering the camera's calibration and metric poses for any unconstrained video would unlock large-scale training data for navigation and scene understanding.
The dominant approaches to this problem are severely limited:
Simultaneous Localization and Mapping (SLAM) is sensitive to initialization and transient failures due to its causal, incremental nature; it is often over-optimized for real-time operation and generally requires known camera calibration;
while Structure-from-Motion (SfM) typically forgoes any image ordering, enabling optimal initialization and global optimization, but lacks robustness to visual symmetries and extreme motions. 
To bridge this gap, we introduce a system that combines the strong sequential constraints of SLAM with the flexibility and global optimization of offline SfM, enabling the metric reconstruction of arbitrary, long, uncalibrated videos.
This system leverages recent advances in wide-baseline dense image matching, treats temporal ordering as a first-class citizen for reliable loop closure, and augments global optimization with metric monocular depth priors.
As a result, thorough evaluations on diverse, challenging datasets that exhibit extreme motion and visual symmetries reveal that our approach is significantly more robust and accurate than both state-of-the-art SLAM and SfM, classical or learned, with given or unknown camera calibration.
The code is publicly available at \url{https://github.com/cvg/vidmap}.
\end{abstract}

\begin{figure*}[t]
  \centering
  \includegraphics[width=\linewidth]{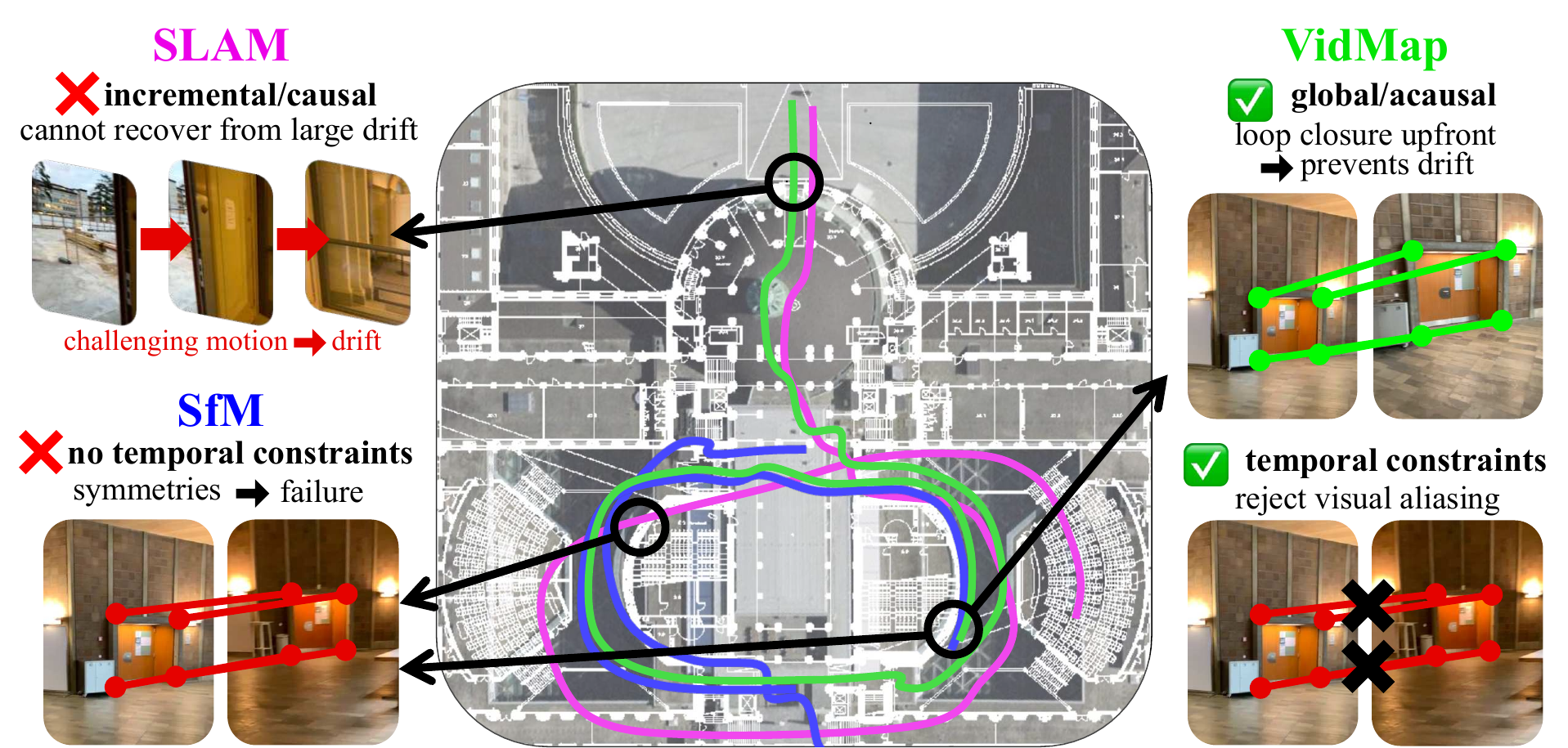}
  \caption{\textbf{\ours\ reconstructs long, unconstrained videos with complex motion and visual ambiguity.}
  SLAM cannot recover from critical tracking failures and drift because it is causal.
  SfM treats videos as unordered image sets and therefore cannot leverage temporal cues to disambiguate visual symmetries.
  \ours\ uses global cues, such as loop closure, upfront to prevent drift, while preserving temporal provenance to reject visual aliasing.
  Moreover, by injecting metric depth constraints into global positioning, VidMap remains robust against scale drift and degenerate motion.
  }
  \label{fig:teaser}
\end{figure*}

\section{Introduction}
\label{sec:intro}

Video is now the dominant medium generated and consumed in the world, with billions of videos available on online platforms.
Robotics and wearable systems also perceive the world through uninterrupted video streams.
To best make use of this data, many computer vision applications require camera poses and calibration.
These cannot be reliably estimated by existing systems, which have been primarily engineered for data captured by expert users and thus cannot handle the challenges of arbitrary videos: degenerate and fast motions, visual symmetries, texture-less environments, or unknown calibration.
These systems typically fall into the categories of SLAM or SfM.

SLAM~\cite{davison2007monoslam,engel2014lsdslam,murartal2015orbslam,campos2021orbslam3,teed2021droidslam,leutenegger2022okvis2,lipson2024dpvslam} is designed for real-time operation and thus registers each new image as it becomes available, in a causal and incremental manner.
The system commits to a single solution at each point in time using only past information, even under partial observability of the environment or motion.
SLAM is thus sensitive to weak initialization, as errors introduced early on are propagated throughout the sequence and compound over time.
Most systems thus require accurate estimates of the camera intrinsics~\cite{murartal2015orbslam,campos2021orbslam3,teed2021droidslam}.
Transient failures, due to degenerate motions, result in a tracking loss that is often unrecoverable.
In practice, academic SLAM systems are optimized for speed and tightly couple all components, making it difficult to explore alternative designs and leverage advances in computer vision.

In contrast, SfM is designed for offline operation and considers all available information before committing to a solution, thus maximizing accuracy.
This process is either global~\cite{moulon2013global,wilson2014robust,pan2024glomap}, by solving for all poses at once, or incremental~\cite{pollefeys2004visual,snavely2006photo,schonberger2016sfm,liu2024hybridsfm}, by registering images in the optimal order, and can jointly optimize the camera calibration.
SfM ignores the sequential structure of videos and can thus likewise handle unordered sets of images~\cite{snavely2006photo,schonberger2016sfm,pan2024glomap}, making it more widely applicable.
This however means that SfM cannot exploit the temporal continuity of videos: it treats all image pairs identically and is thus less robust to visual symmetries.
Additionally, because it performs all data association before motion estimation, SfM is unable to adapt its computation and keyframing to motion dynamics.

In general, classical SfM and SLAM typically lack robustness to textureless environments, poor lighting, blur, and degenerate motion.
Data-driven priors for image matching and tracking~\cite{teed2020raft,sarlin2020superglue,lindenberger2023lightglue,karaev2024cotracker,zhao2022particlesfm}, calibration~\cite{veicht2024geocalib,hagemann2023deep}, metric depth~\cite{piccinelli2024unidepth,wang2024moge,bochkovskiy2025depthpro,hu2024metric3dv2}, 3D surface estimation~\cite{pataki2025mpsfm,duisterhof2024mast3rsfm,murai2024mast3rslam,li2025megasam}, and symmetry disambiguation~\cite{doppelgangers,xiangli2025doppelgangers} improve their robustness but do not address the above-mentioned structural limitations.
Learned end-to-end models~\cite{wang2024dust3r,wang2025vggt} are more robust and efficient but lack precision, scalability, and long-term consistency.

To bridge this gap, we introduce \emph{\ours}, a new system that combines the complementary strengths of SLAM and SfM.
\ours\ leverages the intrinsic temporal ordering of videos to handle extreme motions while avoiding incorrect data association due to visual symmetries.
This ordering makes it easier to leverage dense, pixel-wise matching, which has been historically incompatible with the discrete tracks required by SfM.
Inspired by SLAM, we separate local tracking and long-term loop closure in the global optimization of SfM, thereby maximizing local consistency while reducing long-term drift.
We empower this optimization with metric monocular depth priors to mitigate scale drift and handle degenerate configurations where multi-view constraints alone are insufficient. 

Our system is highly efficient, leveraging motion-based keyframing of SLAM, while being trivially parallelizable over long sequences.

We evaluate \ours\ on new benchmarks~\cite{sarlin2022lamar,Blum_2025_CVPR} that include long videos captured by robots and non-expert users, thus exhibiting extreme motions, viewpoints, and symmetries.
Our experiments reveal that \ours\ outperforms all existing approaches by a large margin, both SLAM and SfM, from classical to end-to-end learned, when camera intrinsics are given or unknown, while retaining similar performance on established saturated benchmarks~\cite{schops2019badslam,burri2016euroc}.

\section{Related Work}
\label{sec:related}

\paragraph{Visual SLAM}
leverages the sequential structure of video by tracking features across consecutive frames, selecting keyframes based on motion, and closing loops to reduce drift.
Classical systems~\cite{campos2021orbslam3,engel2018dso,klein2007ptam} perform well when intrinsics are known and scenes are well-textured but degrade under repetitive structure and require careful calibration.
Learned SLAM methods~\cite{teed2021droidslam,teed2023dpvo,lipson2024dpvslam,czarnowski2020deepfactors,dexheimer2024como} replace hand-crafted components, like tracking, with learned alternatives, improving accuracy in challenging conditions while retaining the same incremental architecture.
Later research integrates dense geometry predictions and monocular depth into SLAM~\cite{murai2024mast3rslam,li2025megasam,huang2025vipe,maggio2025vggtslam,zhang2023goslam}, yet the underlying processing remains causal, estimating the geometry before observing the entire sequence.
This causal commitment is the structural bottleneck: pose and calibration errors from degenerate configurations (forward motion, low parallax, local symmetries) are locked in before well-conditioned observations are used.
Loop closure corrects pose drift but cannot undo decisions already embedded in the map, and tracking failures are unrecoverable.
In contrast, our approach leverages SfM's global optimization and self-calibration to maximize robustness and best use multi-view constraints.

\paragraph{Structure-from-Motion}
optimizes over all observations jointly.
Incremental approaches~\cite{pollefeys2004visual,snavely2006photo,schonberger2016sfm,liu2024hybridsfm} register images one at a time through repeated resectioning and bundle adjustment; global methods~\cite{moulon2013global,wilson2014robust,crandall2011sfm,pan2024glomap} solve for all camera parameters simultaneously.
SfM's optimization is provenance-agnostic, however: it treats all correspondences identically regardless of whether they arise from sequential overlap or loop closure, leaving it vulnerable to visual aliasing from repetitive structure~\cite{doppelgangers,xiangli2025doppelgangers}.
Learned image matching~\cite{detone2018superpoint,sarlin2020superglue,lindenberger2023lightglue,sun2021loftr,edstedt2023dkm,edstedt2024roma} improves the front-end robustness but does not address this structural limitation.
Learned representations have been integrated into SfM~\cite{duisterhof2024mast3rsfm,wang2024vggsfm}, replacing geometric primitives with dense pointmaps or learned tracks but trading geometric precision on long sequences.
ParticleSfM~\cite{zhao2022particlesfm} integrates dense point trajectories from optical flow into global SfM, improving coverage in textureless regions, but constructs correspondences from chained consecutive flow without loop closure, limiting drift correction to local smoothing.
FlowMap~\cite{smith2025flowmap} recovers poses and depth from video via gradient descent over optical flow but is restricted to short sequences without geometric verification.
Methods that integrate monocular depth into classical SfM~\cite{pataki2025mpsfm,liu2022depthsfm} preserve geometric precision but inherit provenance-agnostic matching.
Our approach instead augments global SfM with video-aware tracking and monocular depth priors.
Distinguishing sequential from loop-closure correspondences preserves temporal trust while mitigating visual aliasing; metric depth with per-image scale estimation regularizes degenerate configurations where multi-view constraints alone are insufficient.

\paragraph{Feedforward learned algorithms}

\cite{wang2024dust3r,leroy2024mast3r,wang2025vggt,yang2025da3,wang2026pi3,yang2025fast3r} jointly predict camera poses, calibration, and dense 3D structure from multiple images in a feedforward manner, without any explicit geometric optimization.
Through their learned priors, they achieve impressive robustness in scenarios with sparse views and low visual overlap.
However, their precision remains lower when images have sufficient overlap, and they exhibit significantly more drift and inconsistency with long, high-frequency video sequences.
Since memory and context windows are limited, scaling beyond the training distribution requires chunked alignment~\cite{deng2025vggtlong} or streaming fusion~\cite{maggio2025vggtslam}, re-introducing incremental drift between chunks without propagating uncertainties across them.
Other approaches tailored to videos~\cite{ma2025cut3r,chen2026ttt3r} process frames sequentially with compressed memory, extending effective context through recurrence or test-time training, but degrade beyond their training distribution of motions, environments, and video lengths.
The core tradeoff is structural: learned models gain robustness and speed at the cost of geometric precision, scalability, and zero-shot generalization.
In contrast, we use learned models for what they do well---image matching and metric depth estimation---but rely on explicit geometric optimization for precision and scalability.

\begin{figure*}[t]
  \centering
  \includegraphics[width=\linewidth]{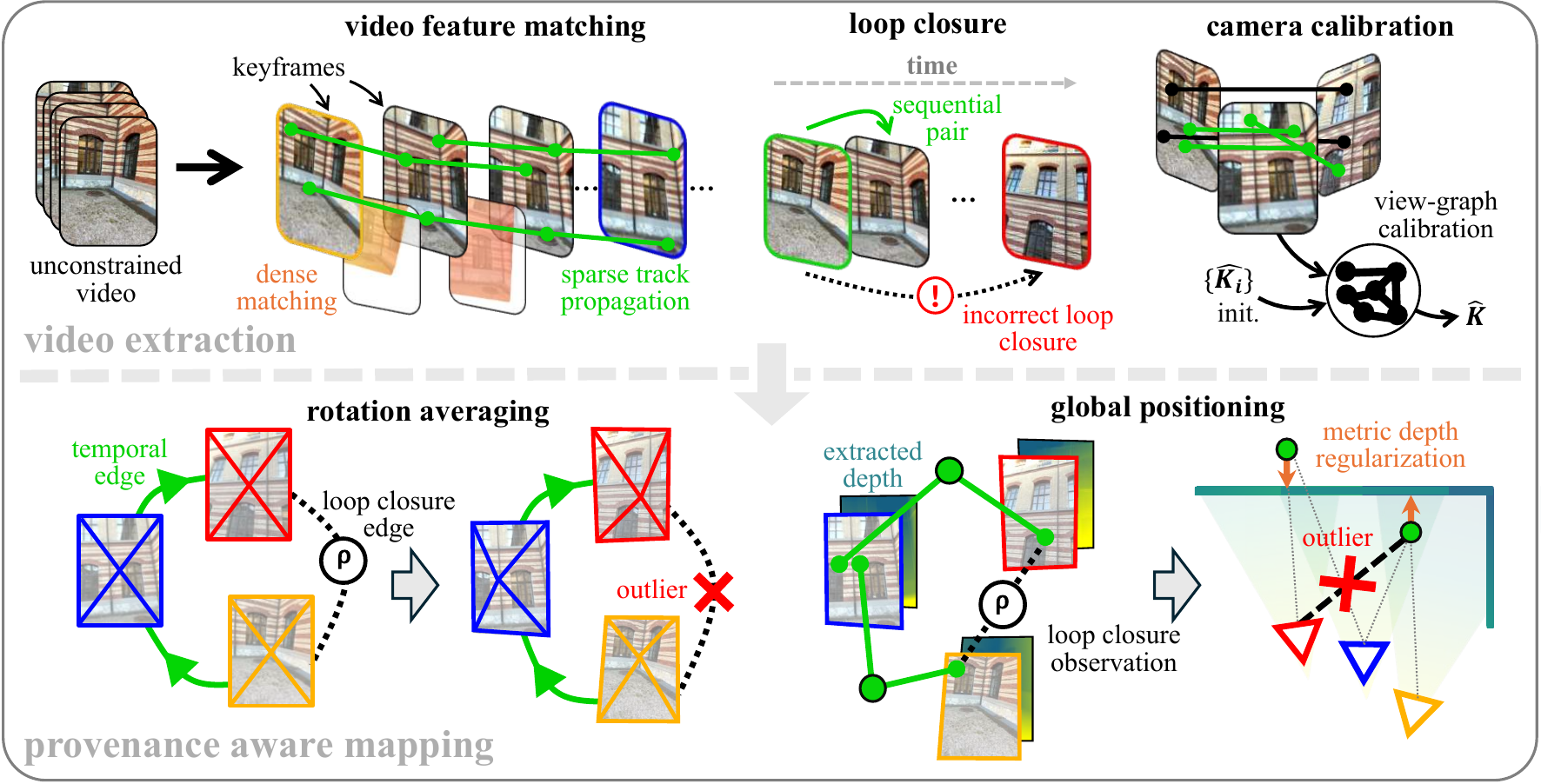}
  \caption{\textbf{System overview.}
    We select keyframes using optical flow and extract monocular depth for each keyframe.
    We sample and propagate sparse tracks across keyframes using dense matching and across loop-closures detected using image retrieval.
    We estimate initial camera intrinsics using view-graph calibration and monocular calibration priors.
    Mapping preserves provenance for pairwise relative poses and track observations: we record which edges are sequential or loop closures to treat them differently.
    Robust optimization therefore treats incorrect loop-closure image pairs as outliers during rotation averaging and rejects loop-closure observations caused by visual aliasing during global positioning.
    By injecting metric depth priors, global positioning remains robust to degenerate motion and scale drift.
  }
  \label{fig:pipeline}
\end{figure*}

\section{Method}
\label{sec:method}

\noindent\textbf{Problem formulation.}
Given a video $\mathbf{I} = \{I_1, \ldots, I_N\}$ with unknown camera intrinsics, we estimate camera poses $T_i \in SE(3)$ for a selected set of keyframes $\mathcal{K} \subset \{1, \ldots, N\}$, intrinsics $K_i$, and a sparse 3D point cloud $\mathbf{X} = \{X_k\}$. 

\paragraph{Overview.}
First, \emph{video-aware extraction} (\cref{sec:matching,sec:loopclosure}) adapts sequential tracking and keyframing concepts from SLAM into an offline dense matching process: selecting keyframes from the continuous video stream, building long sparse tracks with covariance through dense warp chaining, separating loop closure (LC) observations from sequential ones, and extracting monocular depth and focal length priors per keyframe. Although extraction follows the temporal sequence to exploit video structure, it is non-causal: multiple passes revisit the data (low-resolution matching, keyframe detection, high-resolution refinement, track propagation), and no decision is committed until the entire sequence has been observed. Second, \emph{global mapping} (\cref{sec:globalmapping}) introduces new optimization machinery into global SfM~\cite{pan2024glomap}: monocular depth priors with per-image scale estimation in global positioning (GP) and provenance-dependent robust losses that distinguish sequential from LC edges.

\subsection{Video Feature Matching}
\label{sec:matching}

Dense matchers trained on large-scale data have become highly robust to appearance variation, viewpoint change, and textureless surfaces. We leverage this robustness to extract sparse feature tracks with sub-pixel precision and localization covariance, properties that sparse matching cannot provide in low-texture regions. Our approach adapts sequential tracking concepts from SLAM (keyframing, sliding-window propagation, flow chaining) to the dense matching setting, freed from real-time constraints. The key structural property is that consecutive video frames are locally unambiguous: adjacent views share high overlap, and dense matching uses the entire image context to resolve the local ambiguities that confuse sparse or orderless matching. By chaining dense correspondences along time, this guarantee extends across the full video, producing long tracks that are robust to the perceptual aliasing that defeats image retrieval.

\paragraph{Tracking.}
We build on a coarse-to-fine dense matcher~\cite{edstedt2025romav2}. Given images $I_a$ and $I_b$, the matcher yields a dense flow field $\mathcal{W}_{a \to b}: \mathbb{R}^2 \to \mathbb{R}^2$, per-pixel certainty $c_{a \to b} \in [0,1]$, and per-pixel localization covariance $\Sigma_{a \to b} \in \mathbb{R}^{2 \times 2}$.
Unlike classical SfM, which depends on repeatable keypoint detectors, dense matchers predict an accurate flow for each pixel, from which correspondences can be sampled at any location, even in textureless regions. Given sparse keypoints in image $I_a$, each point $\mathbf{x}$ is propagated to $I_b$ via the flow: $\hat{\mathbf{x}}_b = \mathcal{W}_{a \to b}(\mathbf{x}_a)$, with covariance $\Sigma_{a \to b}$ predicted by the matcher. Chaining this propagation across consecutive frames produces long tracks. Under an additive model with independent displacement errors per-pair, the localization covariance progressively grows across sequential pairs of a track as $\Sigma_i^{\text{seq}} = \Sigma_{i-1}^{\text{seq}} + \Sigma_{(i-1) \to i}$.

\paragraph{Keyframe Detection.}
We do not need all the video frames for reconstruction.
We select keyframes adaptively based on observed motion, triggering on tracked image displacement rather than fixed temporal intervals.
For each consecutive frame pair, we estimate the inter-frame motion using the low-resolution flow field.
Starting from sparse keypoints sampled at each keyframe, we propagate them through subsequent frames via the flow field.
We insert a new keyframe when a sufficiently large fraction of keypoints have moved further than a limit distance or lost track \eg, due to occlusion.

\paragraph{Track construction.}
For each selected keyframe $\mathcal{K}_i$, we propagate each track from its location $\mathbf{x}_{i-1}$ in $\mathcal{K}_{i-1}$ to its corresponding location $\mathbf{x}_i$ in $\mathcal{K}_i$. Repeated propagation through consecutive keyframes, however, accumulates tracking drift. Building on~\cite{neoral2024mft}, we reduce this drift by also using direct flows from a window of $W$ preceding keyframes to $\mathcal{K}_i$. We denote the indices of these preceding keyframes by $\mathcal{J}_i := \{k \in \mathbb{N} \mid i-W \leq k < i\}$. For each $j \in \mathcal{J}_i$, applying $\mathcal{W}_{j \to i}$ to the track location $\mathbf{x}_j$ yields the candidate prediction $\hat{\mathbf{x}}_i^{(j)} = \mathcal{W}_{j \to i}(\mathbf{x}_j)$.
For each candidate prediction $\hat{\mathbf{x}}_i^{(j)}$, we compute the propagated localization covariance: $\Sigma_i^{(j)} = \Sigma_j^{\mathrm{seq}} + \Sigma_{j \to i}$.
We select the candidate with minimum covariance trace, indexed by $j^* = \arg\min_{j \in \mathcal{J}_i} \operatorname{tr}(\Sigma_i^{(j)})$.
Predictions from non-adjacent keyframes, however, are more vulnerable to visual aliasing than the sequential prediction $\hat{\mathbf{x}}_i^{\mathrm{seq}} := \hat{\mathbf{x}}_i^{(i-1)}$. We therefore accept the selected candidate only if it agrees sufficiently with the sequential prediction:
\[
\mathbf{x}_i =
\begin{cases}
\hat{\mathbf{x}}_i^{(j^*)}, &
\mathrm{if}\ \left\|\hat{\mathbf{x}}_i^{(j^*)} - \hat{\mathbf{x}}_i^{\mathrm{seq}}\right\|_2^2
< \tau \operatorname{tr}\!\left(\Sigma_i^{(j^*)}\right),\\
\hat{\mathbf{x}}_i^{\mathrm{seq}}, & \text{otherwise}.
\end{cases}
\]
where $\tau$ is a fixed tolerance factor.
\begin{figure*}[t]
  \centering
  \includegraphics[width=\linewidth]{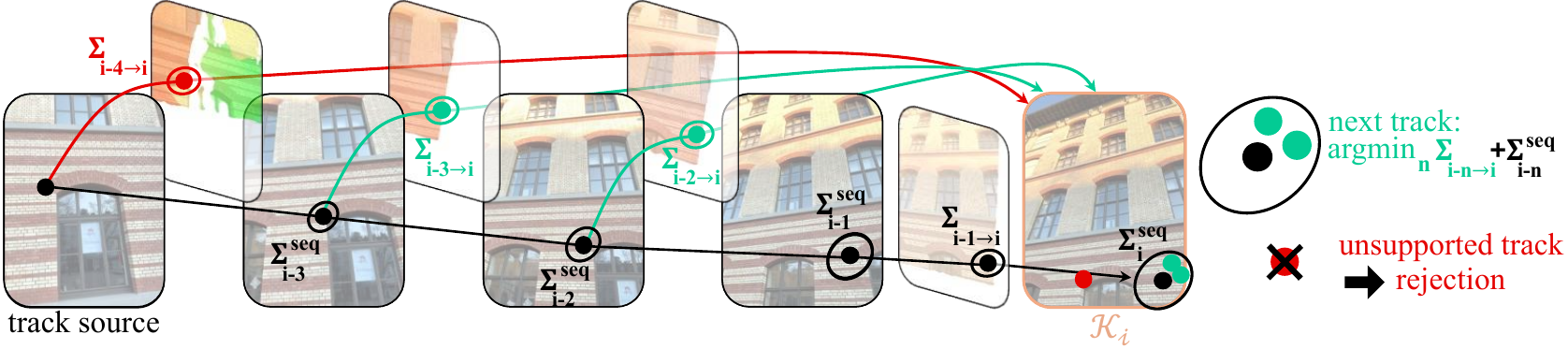}
  \caption{\textbf{Multi-flow drift correction}.
  For each new keyframe $\mathcal{K}_i$, sequential tracks are propagated through the latest match $\mathcal{K}_{i-1}\!\to\!\mathcal{K}_i$, yielding prediction $\hat{\mathbf{x}}_i^{\mathrm{seq}}$ with accumulated covariance $\Sigma_i^{\mathrm{seq}}$. To reduce drift, VidMap also evaluates direct predictions from earlier anchors $\textcolor[RGB]{3,203,148}{\mathcal{K}_{i-n}\!\to\!\mathcal{K}_i}$, yielding $\textcolor[RGB]{3,203,148}{\hat{\mathbf{x}}_i^{(i-n)}}$ with covariance $\textcolor[RGB]{3,203,148}{\Sigma_i^{(i-n)}}$. We select the direct candidate with minimum covariance trace and accept it only if it is consistent with $\hat{\mathbf{x}}_i^{\mathrm{seq}}$; otherwise, $\textcolor{red}{\hat{\mathbf{x}}_i^{(i-n^*)}}$ is rejected and the sequential prediction is retained.
  }
  \label{fig:multiflow}
\end{figure*}

\paragraph{Track Filtering.} We remove tracks that would harm the optimization. Tracks are terminated when their certainty drops below the visibility threshold. For each pair $(\mathcal{K}_{j}, \mathcal{K}_i)$ with $j \in \mathcal{J}_i$, we estimate a fundamental matrix and remove tracks with large epipolar error, catching drift across depth boundaries and hallucinated correspondences through occlusions, which would otherwise introduce systematic errors in triangulation. Finally, we probabilistically thin spatially clustered tracks to maintain uniform coverage (see Appendix~\ref{sec:supp:impl:track-thinning}).

\paragraph{Depth estimation.}
For each selected keyframe $i$, we estimate a monocular depth map~\cite{yang2025da3}, resulting in a depth prior $m_{ik}$ and uncertainty $\sigma_{ik}$ for each observed pixel $k$.
These priors are used during global positioning and bundle adjustment to regularize degenerate geometry and scale drift.

\subsection{Loop Closure}
\label{sec:loopclosure}

Sequential tracking (\cref{sec:matching}) produces reliable tracks across consecutive keyframes but cannot recover constraints between temporally distant views of the same scene. 
Loop closure adds such constraints but may be incorrect and thus corrupt the reconstruction.

\paragraph{Retrieval and Matching.}
We extract global descriptors with MegaLoc~\cite{megaloc} for each keyframe and retrieve the top-$k$ candidates per query that exceed a retrieval similarity threshold, excluding pairs already covered by sequential overlap. Retrieved pairs are matched using the dense matcher at existing keypoint locations and verified through geometric verification.

\begin{figure*}[t]
  \centering
  \includegraphics[width=\linewidth]{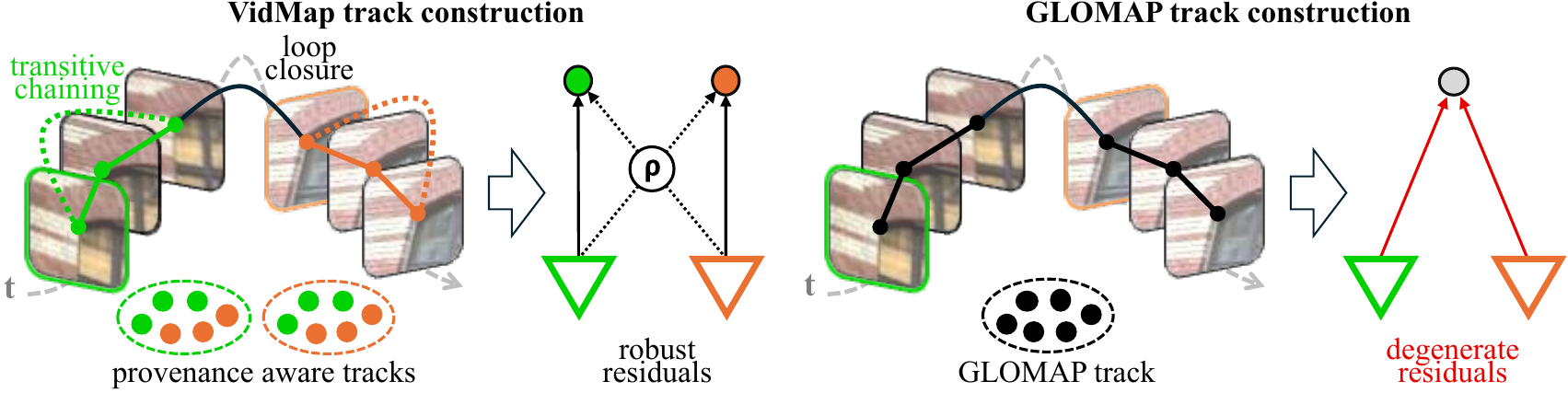}
  \caption{\textbf{Provenance Aware Track Establishment.}
  Left: VidMap constructs tracks from sequential matching and treats loop closure (LC) as a soft link between sequential tracks, which remain reliable when the LC is incorrect.
  Right: GLOMAP merges LC matches by transitivity, so provenance is lost and an incorrect closure can combine separate sequential chains into an inseparable track.
  }
  \label{fig:track_construction}
\end{figure*}

\paragraph{Track Establishment.}
\label{sec:trackestablishment}
Standard global SfM merges all correspondences into tracks via transitivity, making sequential and LC observations indistinguishable. We instead build tracks exclusively from sequential edges. When an LC match links a keypoint that already belongs to a sequential track, we do not merge it via transitivity; instead, the correspondence is attached as a separate observation with provenance label $l_{ik} = \text{lc}$, while sequential observations carry $l_{ik} = \text{seq}$, preserving the chained sequential track. This separation enables provenance-dependent robust losses in global mapping (\cref{sec:globalmapping}): sequential edges are trusted more tightly than LC edges, which may arise from visual aliasing (see \cref{fig:track_construction}).

\begin{figure*}[t]
  \centering
  \includegraphics[width=\linewidth]{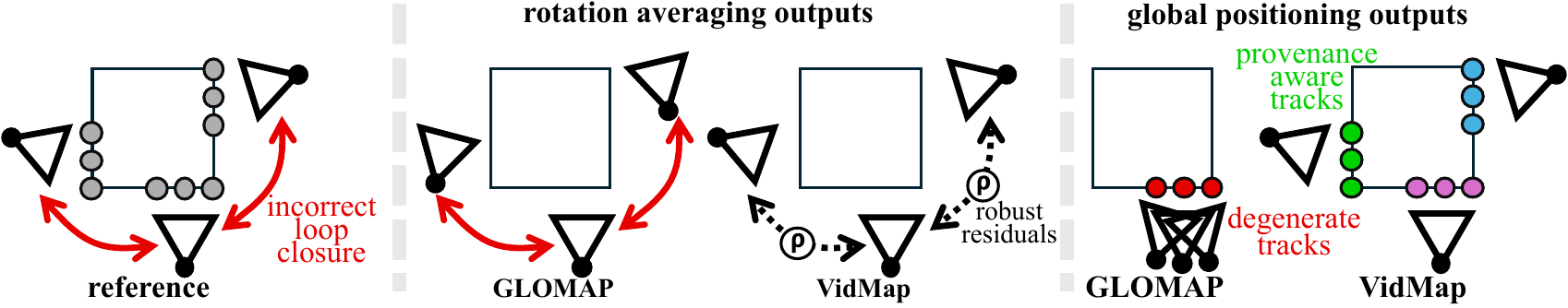}
  \caption{\textbf{Provenance Aware Rotation Averaging and Global Positioning.}
  Under visual symmetry, GLOMAP treats incorrect loop closure (LC) constraints as ordinary edges and observations, corrupting rotation averaging and creating degenerate tracks in global positioning.
  VidMap preserves provenance and robustly downweights LC outliers, recovering the correct orientations and positions.
  }
  \label{fig:ra_gp}
\end{figure*}

\subsection{Uncalibrated Videos}
\label{sec:uncalib}

When camera intrinsics are unknown, we estimate and refine focal lengths in two stages before global mapping.

\paragraph{Focal Length Estimation.}
GeoCalib~\cite{veicht2024geocalib} predicts dense perspective fields per image, from which focal length is recovered via Levenberg-Marquardt optimization.
We first estimate focal length independently per keyframe, then select the top-$k$ most confident keyframes and re-run with a shared-intrinsics constraint, jointly optimizing a single focal length across all $k$ views.

\paragraph{Intrinsics Refinement.} The shared intrinsics are refined via view graph calibration~\cite{sweeney2015viewgraph,pan2024glomap}, which estimates focal lengths from fundamental matrices of well-conditioned pairs, with large baselines and triangulation angles~\cite{schonberger2016sfm}. When insufficient well-conditioned pairs exist, we fall back to the estimate of GeoCalib. The refined intrinsics serve as initialization for bundle adjustment, which jointly optimizes intrinsics alongside poses and structure.

\subsection{Global Mapping}
\label{sec:globalmapping}

We build on the GLOMAP~\cite{pan2024glomap} reconstruction framework, which estimates camera poses through four stages: view graph construction, rotation averaging, GP, and bundle adjustment. We extend this framework with two mechanisms that are new to global SfM. First, monocular depth priors with per-image scale estimation in GP enable metric-scale reconstruction and regularize degeneracies inherent to video sequences where multi-view geometry alone is underconstrained. Second, provenance-dependent robust losses distinguish sequential from LC edges throughout rotation averaging and GP.

\paragraph{Rotation Averaging.}
\label{sec:rotavg}
For each keyframe pair, we estimate pairwise relative rotations $\tilde{R}_{ij}$ based on depth and correspondences~\cite{ding2025reposed}.
Rotation averaging~\cite{pan2024glomap} then estimates global orientations $\{R_i\}$ by minimizing the geodesic distance $\|\log(R_j^\top \tilde{R}_{ij} R_i)\|$ over all edges. We assign provenance-dependent losses: sequential edges receive Huber loss, which trusts the tracked edges, while LC edges receive Cauchy loss, which downweights erroneous closures.

\paragraph{Global Positioning with depth priors.}
\label{sec:gp}
GLOMAP's GP estimates camera centers and 3D points from bearing constraints~\cite{zhuang2018baseline}.
For each observation $(i,k)$, let $m_{ik}$ be the monocular depth prior at the track location and $\sigma_{ik}$ its uncertainty. With the rotations $\{R_i\}$ fixed from the previous step, we jointly optimize camera centers $\{c_i\}$, 3D points $\{X_k\}$, and per-image depth scale factors $\{s_i\}$, where $s_i$ rescales the monocular depth prior for image $i$.
The GP objective combines a \colbearing{bearing residual $\mathbf{e}_{ik}^{\mathrm{GP}}$}, a \coldepth{depth residual $r_{ik}^{\mathrm{depth}}$} that compares the optimized depth to $s_i\cdot m_{ik}$, and a \colscale{scale regularizer $r_i^{\mathrm{scale}}{=}\log s_i$} with scale uncertainty $\sigma_{s,i}$:
\begin{equation}
E_{\text{GP}} = \sum_{(i,k)} \left[
\rho_{l_{ik}}^{\text{GP}}\!\left(
\colbearing{\mathbf{e}_{ik}^{\text{GP}}}^{\top}
{\Sigma_{ik}^{\mathbf{v}}}^{-1}
\colbearing{\mathbf{e}_{ik}^{\text{GP}}}
\right)
+
\rho_{l_{ik}}^{\text{depth}}\!\left(
\frac{\coldepth{r_{ik}^{\text{depth}}}}{\sigma_{ik}}
\right)
\right]
+
\sum_i
\rho^{\text{scale}}\!\left(
\frac{\colscale{r_i^{\text{scale}}}}{\sigma_{s,i}}
\right)
\label{eq:gp}
\end{equation}
Each observation carries provenance label $l_{ik} \in \{\text{seq}, \text{lc}\}$ (\cref{sec:loopclosure}), and the robust loss $\rho_{l_{ik}}$ is selected by provenance. 
We now define the residuals.

\paragraph{Bearing residual.}
This penalizes the angular discrepancy between the observed camera ray $\mathbf{v}_{ik}$ and the direction to the 3D point:
\begin{equation}
\colbearing{\mathbf{e}_{ik}^{\text{GP}}} = \mathbf{v}_{ik} - d_{ik}(X_k - c_i)
\label{eq:geomresidual}
\end{equation}
where $\mathbf{v}_{ik}$ is the globally rotated camera ray from $c_i$ through observation $\mathbf{x}_{ik}$ and $d_{ik} \geq 0$ is a normalizing factor. In BA, where rotations are free, this is replaced by the standard reprojection error $\mathbf{e}_{ik}^{\text{BA}} = \pi(K_i, T_i, X_k) - \mathbf{x}_{ik}$ in pixel space.

\paragraph{Covariance weighting.}
Each observation carries a 2D pixel covariance $\Sigma_{ik} \in \mathbb{R}^{2 \times 2}$ from track extraction. In BA, this weights the reprojection error directly. In GP, the residual lives in 3D bearing space. We propagate $\Sigma_{ik}$ through the unprojection and rotation into a 3D bearing covariance $\Sigma_{ik}^{\mathbf{v}} = J_i\, \Sigma_{ik}\, J_i^\top \in \mathbb{R}^{3 \times 3}$, where $J_i$ is the Jacobian of the camera-frame bearing with respect to pixel coordinates. We provide the full derivation in the supplementary.

\paragraph{Depth residual.}
For each observation $(i,k)$, the camera-frame depth $z_{ik}$ should agree with the monocular prediction $m_{ik}$ after accounting for the per-image scale $s_i$. Points in front of the camera use a log-space residual; points behind the camera fall back to linear:
\begin{equation}
\coldepth{r_{ik}^{\text{depth}}} =
\begin{cases}
\log\!\left(\dfrac{z_{ik}}{s_i \cdot m_{ik}}\right) & z_{ik} > 0 \\[6pt]
z_{ik} - s_i \cdot m_{ik} & z_{ik} \leq 0
\end{cases}
\quad \mathrm{where}\quad
z_{ik} = \mathbf{e}_3^\top R_i(X_k - c_i)
\label{eq:depthresidual}
\end{equation}
\subsubsection{Bundle Adjustment.}
\label{sec:ba}

This refines poses $\{T_i\}$, 3D points $\{X_k\}$, intrinsics $\{K_i\}$, and per-image depth scales $\{s_i\}$. In contrast to MP-SfM~\cite{pataki2025mpsfm}, which fixes the depth scales after incremental reconstruction, we also optimize them:
\begin{equation}
E_{\text{BA}} =
\sum_{(i,k)} \left[
\rho^{\text{BA}}\!\left(
\mathbf{e}_{ik}^{\text{BA}\top}
\Sigma_{ik}^{-1}
\mathbf{e}_{ik}^{\text{BA}}
\right)
+
\rho^{\text{depth}}\!\left(
\frac{\coldepth{r_{ik}^{\text{depth}}}}{\sigma_{ik}}
\right)
\right]
+
\sum_i
\rho^{\text{scale}}\!\left(
\frac{\colscale{r_i^{\text{scale}}}}{\sigma_{s,i}}
\right)
\label{eq:ba}
\end{equation}
where $\mathbf{e}_{ik}^{\text{BA}} = \pi(K_i, T_i, X_k) - \mathbf{x}_{ik}$ is the reprojection error in pixels, weighted by the 2D pixel covariance $\Sigma_{ik}$ from track extraction. By the BA stage, the reconstruction is near the correct solution and geometric verification has already filtered inconsistent loop closure observations, so we apply uniform robust losses to all observations regardless of provenance.

\paragraph{Robust Optimization.}
Inspired by graduated non-convexity~\cite{yang2020gnc}, we progressively tighten the robust losses across iterations in GP and BA.
Following MP-SfM~\cite{pataki2025mpsfm}, the Cauchy scale for depth residuals is annealed as the geometry stabilizes.
We additionally flag depth-match inconsistencies by projecting monocular depth through the pairwise relative pose and switch the corresponding depth residuals to Cauchy loss when the depth ratio exceeds a threshold.

\begin{table}[tb]
  \caption{\textbf{Evaluation on the LaMAR dataset~\cite{sarlin2022lamar}.}
  We report the AUC of the translation error (in \%, higher is better) for windows of different lengths.
  \ours\ outperforms all existing approaches in both calibrated and uncalibrated settings.
  }
  \label{tab:lamar_wte_auc}
  \label{tab:lamar}
  \centering
  \setlength{\tabcolsep}{3pt}
  \begin{tabular}{l ccccc ccccc}
    \toprule
    & \multicolumn{5}{c}{uncalibrated} & \multicolumn{5}{c}{calibrated} \\
    \cmidrule(lr){2-6} \cmidrule(lr){7-11}
    \multicolumn{1}{r@{\hskip 8pt}}{W-AUC $\to$} & 10\,m & 25\,m & 50\,m & 100\,m & full & 10\,m & 25\,m & 50\,m & 100\,m & full \\
    \midrule
    LoGeR & 65.9 & 64.5 & 60.3 & 60.1 & 59.8 & \multicolumn{5}{c}{\cinvalid{cannot use calibration}} \\
    VGGT-SLAM2 & 74.0 & 69.2 & 60.3 & 49.4 & 56.9 & \multicolumn{5}{c}{\cinvalid{cannot use calibration}} \\
    Lingbot-Map & 69.2 & 75.6 & 75.0 & \cthird 72.7 & 69.62 & \multicolumn{5}{c}{\cinvalid{cannot use calibration}} \\
    DA3-Long & \csecond 86.7 & \csecond 86.7 & \csecond 84.6 & \csecond 78.1 & \cthird 76.5 & \multicolumn{5}{c}{\cinvalid{cannot use calibration}} \\
    DPV-SLAM & \multicolumn{5}{c}{\cinvalid{requires calibration}} & 53.1 & 39.0 & 25.1 & 14.9 & 38.2 \\
    DROID-W & \multicolumn{5}{c}{\cinvalid{requires calibration}} & \cthird 88.0 & \cthird 84.7 & \cthird 80.2 & \csecond 79.5 & \cthird 76.2 \\
    MP-SfM & \multicolumn{5}{c}{\cinvalid{requires calibration}} & \csecond 88.6 & 83.7 & 74.5 & \cthird 60.2 & 60.6 \\
    GLOMAP-LG & 76.4 & 69.4 & 60.1 & 53.6 & 62.3 & 77.3 & 70.4 & 61.6 & 48.4 & 63.0 \\
    GLOMAP-RoMA & 71.0 & 66.0 & 58.4 & 46.7 & 59.4 & 75.7 & 69.0 & 59.2 & 42.1 & 60.1 \\
    MASt3R-SLAM & 60.8 & 63.6 & 55.6 & 47.3 & 51.5 & 69.3 & 68.8 & 65.0 & 58.1 & 57.7 \\
    MegaSaM & 77.5 & 68.4 & 59.9 & 51.4 & 52.7 & 78.1 & 69.2 & 58.3 & 52.1 & 55.6 \\
    ViPE & \cthird 86.6 & \cthird 83.8 & \cthird 80.2 & \csecond 78.1 & \csecond 76.6 & 87.3 & \csecond 84.9 & \csecond 80.4 & \csecond 79.5 & \csecond 78.8 \\
    VidMap & \cfirst 91.9 & \cfirst 92.3 & \cfirst 89.9 & \cfirst 89.3 & \cfirst 88.5 & \cfirst 93.2 & \cfirst 93.0 & \cfirst 90.5 & \cfirst 89.9 & \cfirst 89.7 \\
    \bottomrule
  \end{tabular}
\end{table}

\section{Experiments}
\label{sec:experiments}

\subsection{Datasets}
\label{sec:datasets}
We evaluate the accuracy of video camera pose estimation using four datasets.
All hyperparameters used throughout the system are fixed across all datasets and were tuned on separate validation sequences.

\paragraph{LaMAR}~\cite{sarlin2022lamar}
is a collection of videos recorded by phones and AR devices in three large indoor scenes (CAB, HGE, LIN), with ground truth poses and calibration.
We select 50 phone videos from those in the mapping split, whose ground truth poses are publicly available.
This dataset exhibits long trajectories with complex motion, visual aliasing, textureless surfaces, and degenerate viewpoints, which all stress both data association and trajectory optimization.

\paragraph{CroCoDL}~\cite{Blum_2025_CVPR}
is a collection of sequences recorded by phones (62) and robots (26) in 4 disaster-site buildings, with images featuring complex motion and unstructured built environments.
This dataset is suitable for evaluating the generalization of learned approaches outside of their training distribution.

\paragraph{ETH3D-SLAM}~\cite{schops2019badslam}
contains 55 indoor and outdoor sequences recorded with a hand-held camera, featuring high-accuracy ground truth poses and calibration and varying scene complexity and camera motion.

\paragraph{EuRoC}~\cite{burri2016euroc}
contains 11 sequences captured by a drone, featuring grayscale stereo fisheye cameras and fast 6-DOF motion with blur, which are both challenging for learned computer vision models.

\subsection{Baselines and Metrics}
\label{sec:baselines}

\paragraph{Baselines.}
We evaluate existing approaches that belong to three categories.
\emph{SLAM and visual odometry:}
DPV-SLAM~\cite{lipson2024dpvslam} is based on learned patch tracking and refinement;
MegaSaM~\cite{li2025megasam} and ViPE~\cite{huang2025vipe} combine learned optical flow with off-the-shelf monocular depth and calibration priors;
DROID-W~\cite{li2026droidw} also masks out dynamic objects;
MASt3R-SLAM~\cite{murai2024mast3rslam} relies on two-view prediction of dense pointmaps;
\emph{End-to-end deep models:}
DA3-Long~\cite{yang2025da3,deng2025vggtlong} aligns overlapping subsequences reconstructed using Depth Anything 3~\cite{yang2025da3}.
LoGeR~\cite{zhang2026loger} and LingBot-Map~\cite{chen2026lingbotmap} leverage long-context mechanisms.
\emph{Global SfM:} 
We evaluate GLOMAP~\cite{pan2024glomap} with SuperPoint+LightGlue~\cite{detone2018superpoint,lindenberger2023lightglue} and SuperPoint+RoMA v2~\cite{detone2018superpoint,edstedt2025romav2}.
We evaluate two calibration settings: \emph{calibrated} uses ground-truth intrinsics; \emph{uncalibrated} estimates intrinsics from the video without prior.

\paragraph{Metrics.}
For ETH3D and EuRoC, we report the area under the recall curve (AUC, \%) of the translation error after global alignment.
Since the alignment is easily dominated by drift and misses fine-grained local accuracy, we proceed differently for LaMAR and CroCoDL, which exhibit longer sequences:
we compute a windowed AUC (W-AUC, \%) over local windows of each trajectory, in which we adapt the error threshold to 5\% of the window size -- see details in the supplementary material.

\begin{table}[tb]
  \caption{\textbf{Evaluation on the CroCoDL dataset~\cite{Blum_2025_CVPR}}.
  We report the AUC of the translation error (in \%, higher is better) for windows of different lengths.
  \ours\ outperforms all existing approaches by a large margin, especially on the challenging robot sequences, in both calibrated and uncalibrated settings.
  }
  \label{tab:crocodl_wte_auc}
  \label{tab:crocodl}
  \centering
  \setlength{\tabcolsep}{2pt}
  \begin{tabular}{c l ccccccccc}
    \toprule
    & & \multicolumn{5}{c}{Phone} & \multicolumn{4}{c}{Robot} \\
    \cmidrule(lr){3-7} \cmidrule(lr){8-11}
    & \multicolumn{1}{r@{\hskip 8pt}}{Window-AUC $\to$} & 10\,m & 25\,m & 50\,m & 100\,m & full & 10\,m & 25\,m & 50\,m & full \\
    \midrule
    \multirow{10}{*}{\rotatebox{90}{uncalib.}} & VGGT-SLAM2 & 62.8 & 64.3 & 66.0 & 62.5 & 74.9 & 34.1 & 31.9 & 33.1 & 52.8 \\
    & Lingbot-Map & 54.4 & 66.3 & 69.9 & 72.2 & 72.0 & 60.9 & 40.6 & 63.0 & 61.6 \\
    & LoGeR & 50.0 & 56.7 & 57.9 & 63.7 & 63.8 & 47.4 & 55.8 & 62.3 & 65.5 \\
    & DA3-Long & \csecond 86.2 & \csecond 88.1 & \csecond 88.5 & \csecond 88.8 & \csecond 88.9 & \csecond 79.5 & \csecond 82.4 & \csecond 82.0 & \csecond 78.3 \\
    & GLOMAP-LG & \cthird 77.8 & 66.6 & 60.0 & 65.9 & \cthird 76.4 & 10.2 & 7.4 & 10.3 & 19.2 \\
    & GLOMAP-RoMA & 66.2 & 57.8 & 53.5 & 56.3 & 74.8 & 12.7 & 10.0 & 10.3 & 20.3 \\
    & MASt3R-SLAM & 51.8 & 57.3 & 60.2 & 65.2 & 70.0 & 25.8 & 30.4 & 32.2 & 46.1 \\
    & MegaSaM & 50.2 & 50.2 & 43.5 & 35.2 & 50.1 & 41.9 & 39.5 & 32.4 & 45.5 \\
    & ViPE & 66.7 & \cthird 68.2 & \cthird 68.6 & \cthird 70.7 & 71.4 & \cthird 59.5 & \cthird 61.4 & \cthird 66.2 & \cthird 71.0 \\
    & VidMap & \cfirst 94.0 & \cfirst 95.0 & \cfirst 95.3 & \cfirst 93.0 & \cfirst 95.5 & \cfirst 91.4 & \cfirst 85.7 & \cfirst 82.5 & \cfirst 80.3 \\
    \midrule
    \multirow{8}{*}{\rotatebox{90}{calib.}} & DPV-SLAM & 19.8 & 15.6 & 14.9 & 19.7 & 31.8 & 66.1 & \cthird 68.3 & \csecond 73.0 & \csecond 77.0 \\
    & DROID-W & 71.8 & \cthird 70.4 & \cthird 67.9 & \cthird 70.6 & 72.7 & \cthird 71.9 & \csecond 72.7 & \cthird 71.0 & 67.8 \\
    & GLOMAP-LG & \cthird 73.9 & 66.8 & 62.0 & 57.9 & 72.2 & 42.7 & 38.9 & 34.1 & 42.4 \\
    & GLOMAP-RoMA & \csecond 77.3 & 65.2 & 57.2 & 60.6 & \csecond 75.3 & 45.9 & 42.5 & 37.7 & 45.0 \\
    & MASt3R-SLAM & 50.1 & 52.6 & 43.8 & 38.6 & 63.6 & 29.8 & 29.5 & 27.1 & 45.4 \\
    & MegaSaM & 68.1 & 59.8 & 51.0 & 50.3 & 52.3 & \csecond 74.5 & 59.8 & 43.0 & 45.7 \\
    & ViPE & 70.0 & \csecond 71.6 & \csecond 71.0 & \csecond 71.1 & \cthird 74.3 & 62.1 & 64.5 & 68.9 & \cthird 72.2 \\
    & VidMap & \cfirst 94.7 & \cfirst 95.8 & \cfirst 96.3 & \cfirst 95.6 & \cfirst 96.6 & \cfirst 91.5 & \cfirst 85.2 & \cfirst 80.6 & \cfirst 78.2 \\
    \bottomrule
  \end{tabular}
\end{table}

\subsection{Results}

\begin{figure}[p]
    \centering
    \includegraphics[width=0.94\linewidth]{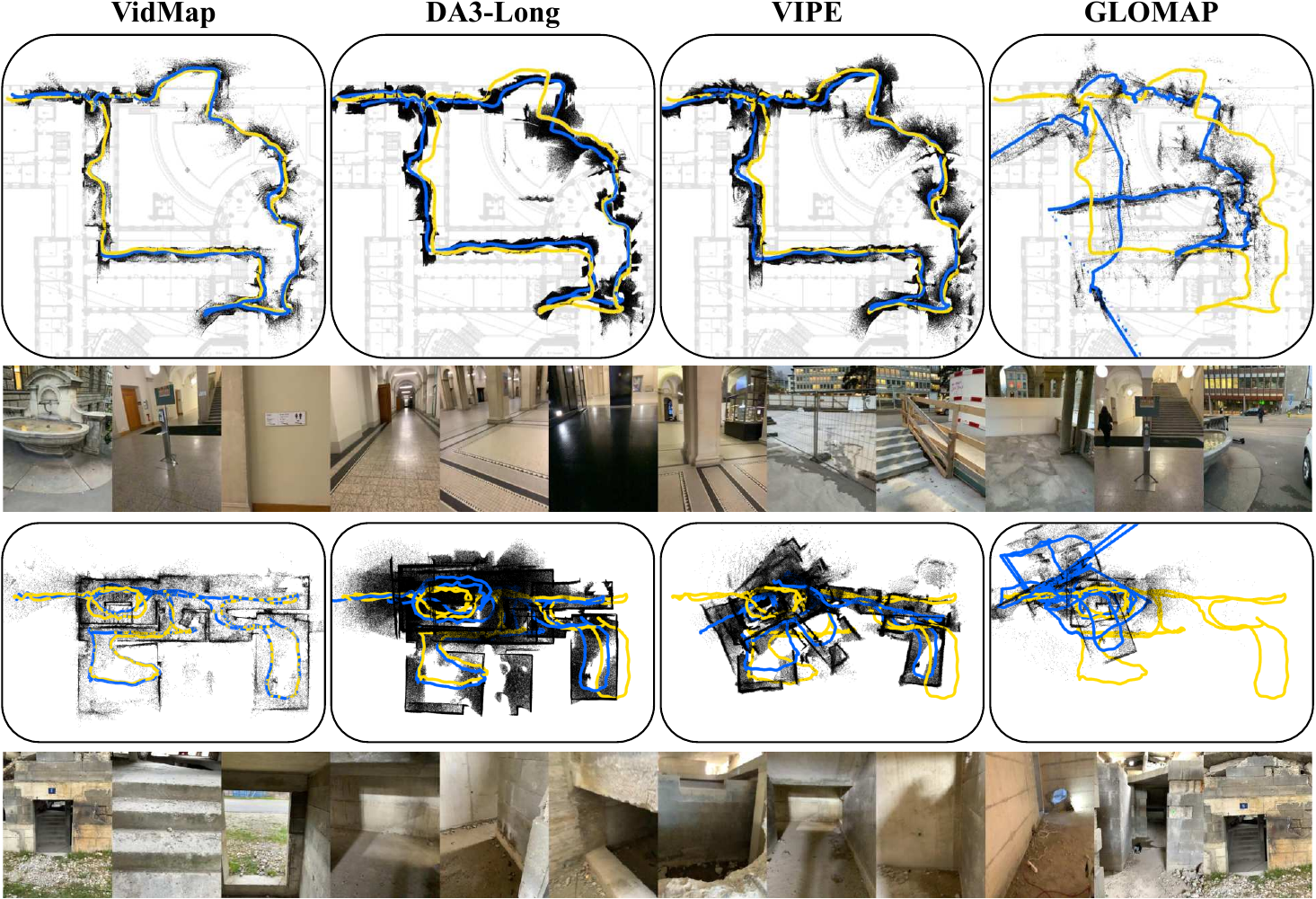}
    \caption{\textbf{Qualitative comparison to existing approaches on LaMAR (top) and CroCoDL (bottom).}
Each row compares \ours\ with DA3-Long, ViPE, and GLOMAP on one scene and shows representative input images.
\ours\ reconstructs the scenes without drift, while others suffer from drift or lose track.}
    \label{fig:baseline_comparison}
    \vspace{0.1cm}
    \includegraphics[width=0.94\textwidth]{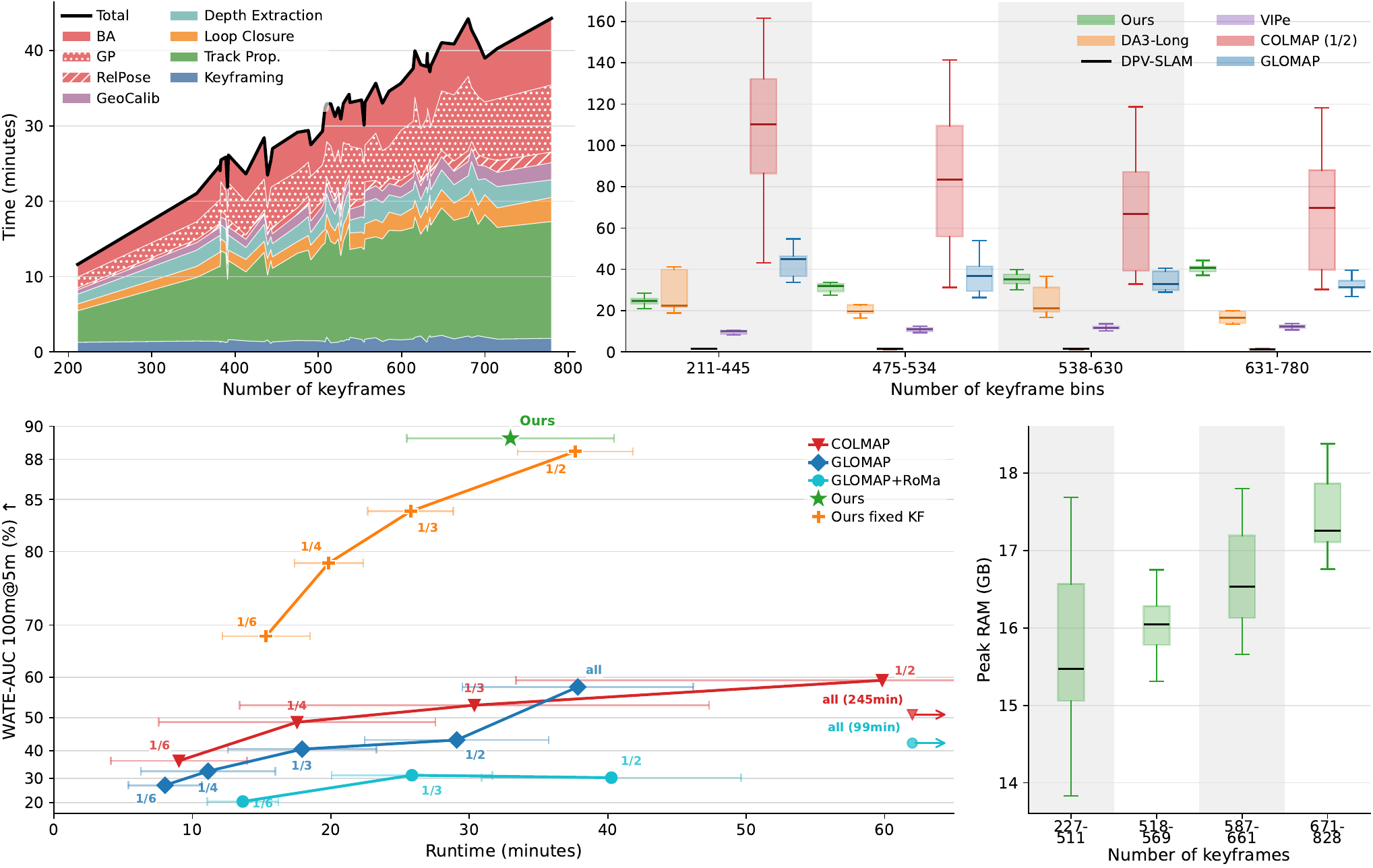}
    \caption{\textbf{Analysis of the runtime.}
    Top left: Duration of each component with increasing number of keyframes. Matching is the most expensive step.
    Top right: Comparison of the total runtime against existing approaches.
    Bottom left: Impact of keyframing on the speed and accuracy. Motion-aware keyframing retains the accuracy of the dense keyframing but at lower cost, while aggressive keyframing reduces runtime at the cost of accuracy.
    Bottom right: peak RAM usage remains practical with longer videos.}
    \label{fig:runtime}
\end{figure}

\vspace{-0.2cm}
\paragraph{LaMAR -- \cref{tab:lamar}, \cref{fig:baseline_comparison}.}
Long sequences exhibit geometric degeneracies, such as narrow baselines, forward motion, or textureless surfaces, that are challenging for classical approaches:
GLOMAP and DPV-SLAM, which rely on multi-view geometry alone, drift by an order of magnitude more than \ours.
Approaches that leverage metric depth priors, such as ViPE and MegaSaM, are more robust to these degeneracies and maintain limited drift, but still fall short on precision.
\ours\ also outperforms all existing approaches when given ground-truth calibration, confirming that no single component but rather the interplay of video-aware extraction and depth-augmented global optimization closes the gap.
The accuracy of VidMap is nearly identical in the uncalibrated and GT-calibrated settings, confirming that VidMap estimates accurate calibration.

\paragraph{CroCoDL -- \cref{tab:crocodl}, \cref{fig:baseline_comparison}.} CroCoDL evaluates generalization to out-of-domain videos captured in disaster-site buildings. DA3-Long, whose depth model is trained on standard indoor/outdoor scenes, degrades substantially on this out-of-distribution data. GLOMAP's pure geometry is limited by the lack of video-aware extraction. \ours\ outperforms all baselines by a large margin on both phone and robot sequences, without retraining or domain-specific adaptation.

\paragraph{ETH3D-SLAM and EuRoC -- \cref{tab:eth3d_euroc}.} 
Because these sequences are shorter and well-conditioned, multi-view geometry alone is sufficient and geometric priors primarily add noise, making DPV-SLAM the strongest baseline.
\ours\ outperforms it by a significant margin across all thresholds on ETH3D and performs on par on EuRoC,
which features grayscale fisheye cameras and motion blur, making learned depth and calibration priors less reliable and helpful.

In the uncalibrated setting, 
\ours\ significantly outperforms all baselines, classical and end-to-end learned,
with a larger gap to the closest baselines ViPE and MegaSaM.
The 18-point gap between VidMap's calibrated and uncalibrated results confirms that calibration remains the primary bottleneck for precision when the motion is easier.

\begin{table}[tb]
  \caption{\textbf{Evaluation on the ETH3D-SLAM and EuRoC datasets.}
  We report AUC of the translation error (in \%, higher is better) for different error thresholds.
  }
  \label{tab:eth3d_euroc_pose_auc}
  \label{tab:eth3d_euroc}
  \centering
  \setlength{\tabcolsep}{2pt}
  \begin{tabular}{c l ccccc ccccc}
    \toprule
    & & \multicolumn{5}{c}{ETH3D-SLAM} & \multicolumn{5}{c}{EuRoC} \\
    \cmidrule(lr){3-7} \cmidrule(lr){8-12}
    & \multicolumn{1}{r@{\hskip 8pt}}{AUC@Xm $\to$} & 5\,cm & 10\,cm & 50\,cm & 1\,m & 10\,m & 5\,cm & 10\,cm & 50\,cm & 1\,m & 10\,m \\
    \midrule
    \multirow{10}{*}{\rotatebox{90}{uncalibrated}} & GLOMAP-LG & 20.4 & 30.9 & 58.7 & 67.4 & 77.8 & \00.4 & \01.8 & 16.1 & 29.7 & 70.7 \\
    & GLOMAP-RoMA & 27.7 & 39.3 & 56.8 & 60.8 & 65.5 & \00.0 & \00.2 & 12.3 & 27.9 & 64.8 \\
    & LoGeR & \06.4 & 16.9 & 57.7 & 73.8 & 95.9 & \00.1 & \00.5 & 18.2 & 45.1 & 92.6 \\
    & Lingbot-Map & 13.1 & 27.8 & 68.3 & 81.3 & 97.2 & \00.1 & \00.7 & 32.0 & 55.5 & 91.7 \\
    & DA3-Long & 19.0 & 37.2 & 73.6 & 84.4 & \csecond 98.0 & \00.1 & \00.8 & 31.4 & 57.6 & 95.4 \\
    & VGGT-SLAM2 & 26.4 & 44.1 & 75.7 & 84.3 & 97.1 & \cthird \01.3 & \csecond \07.8 & \csecond 55.4 & \cthird 74.2 & 96.9 \\
    & MASt3R-SLAM & \08.9 & 21.6 & 65.3 & 77.0 & 96.4 & \csecond \01.6 & \05.6 & \cthird 51.0 & 71.9 & \cthird 97.0 \\
    & MegaSaM & \cthird 38.4 & \cthird 53.9 & \csecond 79.2 & \csecond 86.7 & \csecond 98.0 & \00.8 & \05.7 & 50.3 & 73.2 & \csecond 97.3 \\
    & ViPE & \csecond 38.7 & \csecond 54.9 & \cthird 78.3 & \cthird 85.7 & \cthird 97.6 & \cthird \01.3 & \cthird \06.2 & \csecond 55.4 & \csecond 75.0 & 94.0 \\
    & VidMap & \cfirst 51.4 & \cfirst 68.3 & \cfirst 90.1 & \cfirst 94.0 & \cfirst 99.0 & \cfirst 11.5 & \cfirst 31.6 & \cfirst 79.1 & \cfirst 89.4 & \cfirst 99.0 \\
    \midrule
    \multirow{8}{*}{\rotatebox{90}{calibrated}} & GLOMAP-LG & 27.9 & 38.6 & 64.6 & 72.3 & 80.9 & 12.8 & 35.3 & 72.4 & 79.8 & 91.8 \\
    & GLOMAP-RoMA & 37.8 & 47.2 & 59.6 & 63.0 & 67.0 & 18.0 & 43.2 & 79.0 & 85.1 & 90.9 \\
    & MASt3R-SLAM & 22.7 & 37.5 & 70.2 & 78.5 & 96.8 & \05.6 & 18.3 & 68.1 & 83.2 & 98.3 \\
    & MegaSaM & 53.8 & 62.9 & 80.4 & 87.4 & 98.0 & 15.6 & 35.4 & 84.1 & 92.1 & \cthird 99.2 \\
    & ViPE & 54.4 & 64.9 & 82.2 & 88.4 & 98.2 & \05.5 & 26.4 & 79.9 & 89.9 & 99.0 \\
    & DROID-W & \csecond 63.8 & \cthird 72.4 & \cthird 86.2 & \cthird 91.0 & \csecond 98.6 & \cthird 20.9 & \cthird 51.0 & \cthird 90.1 & \cthird 95.0 & \csecond 99.5 \\
    & DPV-SLAM & \cthird 62.3 & \csecond 72.5 & \csecond 87.4 & \csecond 91.8 & \cthird 98.4 & \csecond 22.5 & \csecond 53.5 & \csecond 90.6 & \csecond 95.3 & \csecond 99.5 \\
    & VidMap & \cfirst 69.4 & \cfirst 79.2 & \cfirst 92.2 & \cfirst 95.0 & \cfirst 99.1 & \cfirst 23.0 & \cfirst 54.6 & \cfirst 90.9 & \cfirst 95.4 & \cfirst 99.6 \\
    \bottomrule
  \end{tabular}
\end{table}

\begin{table}[tb]
  \caption{%
  \textbf{Ablation study on the LaMAR and ETH3D-SLAM datasets.}
   Metric depth priors are most critical for long-range accuracy, while provenance-aware losses and loop closure improve robustness to incorrect multi-view observations.
  }
  \label{tab:ablations}
  \centering
  \setlength{\tabcolsep}{2pt}
  \footnotesize
  \begin{tabular}{l ccccc ccccc}
    \toprule
    & \multicolumn{5}{c}{LaMAR (Window-AUC $\uparrow$)} & \multicolumn{5}{c}{ETH3D-SLAM (AUC $\uparrow$)} \\
    \cmidrule(lr){2-6} \cmidrule(lr){7-11}
    & 10\,m & 25\,m & 50\,m & 100\,m & full & 5\,cm & 10\,cm & 50\,cm & 1\,m & 10\,m \\
    \midrule
    \textbf{VidMap} & \cfirst 91.9 & \cfirst 92.3 & \cfirst 89.9 & \cfirst 89.3 & \cfirst 88.5 & \csecond 51.4 & \csecond 68.3 & \cfirst 90.1 & \cfirst 94.0 & \cfirst 99.0 \\
    \midrule
    \multicolumn{11}{l}{\emph{Depth priors}} \\
    No depth in GP & 67.6 & 42.8 & 24.9 & 13.6 & 37.7 & 50.5 & \cthird 66.5 & \cthird 87.3 & \cthird 91.9 & \cthird 98.7 \\
    No depth in BA & \csecond 91.5 & \csecond 91.1 & \csecond 88.0 & \cthird 86.5 & \csecond 85.5 & 44.5 & 60.6 & 86.0 & \cthird 91.9 & \csecond 98.8 \\
    No scale optimization & 69.3 & 44.7 & 25.6 & 13.9 & 39.7 & 50.2 & 65.9 & \csecond 87.4 & \csecond 92.4 & \csecond 98.8 \\
    No metric scale & 90.2 & 88.0 & 81.9 & 71.7 & 74.1 & \cfirst 52.2 & \cfirst 68.8 & \cfirst 90.1 & \cfirst 94.0 & \cfirst 99.0 \\
    \midrule
    \multicolumn{11}{l}{\emph{Temporal structure}} \\
    No loop closure edges & \cthird 90.8 & \cthird 90.3 & \cthird 87.6 & \csecond 86.6 & \csecond 85.5 & 32.9 & 47.4 & 77.5 & 85.9 & 97.9 \\
    No provenance losses & 88.4 & 85.7 & 80.2 & 77.0 & 79.9 & \cthird 50.6 & 66.1 & 87.1 & 91.5 & 98.4 \\
    \bottomrule
  \end{tabular}
\end{table}

\subsection{Detailed Analysis}
\label{sec:ablations}
\vspace{-0.2cm}

\paragraph{Ablation study.}
\Cref{tab:ablations} measures the impact of key design decisions and shows that all components of \ours\ are critical to its high robustness and accuracy.

Removing depth priors in either GP or BA incurs a drop in accuracy because these priors add geometric constraints that prevent degenerate estimates under complex motion.
\emph{No scale optimization} relies on the monocular metric scale and prevents per-image refinement.
\emph{No metric scale} discards the metric prior, optimizing the per-image scale solely based on multi-view constraints.
Both variants impair the robustness on LaMAR over longer horizons.
\ours\ thus uses metric monocular depth as a soft prior without rigid constraint.
Overall, the benefits of depth and metric scale are smaller on ETH3D-SLAM, where sequences are shorter and motion is better constrained.
Removing loop closure edges, \ie relying only on sequential constraints, incurs a large drift on ETH3D-SLAM.
Treating sequential and loop closure edges equally (\emph{no provenance losses}) is prone to visual aliasing and results in catastrophic failure on LaMAR.

\paragraph{Runtime.}
We analyze the speed and memory requirements of \ours\ in \cref{fig:runtime}.
\ours\ is an offline system but can efficiently handle long videos, as
time grows linearly with keyframes and the required memory remains practical.
\ours\ is comparably as efficient as GLOMAP and substantially faster than COLMAP.

\section{Conclusion}
\label{sec:conclusion}
In this paper, we introduced \ours, a reconstruction framework that bridges the fundamental divide between the temporal awareness of causal SLAM and the robust global optimization of offline SfM.
By explicitly distinguishing sequential tracking from loop-closure observations and integrating metric depth priors into global optimization, we formulate a non-causal paradigm that preserves temporal trust without prematurely committing to an incremental geometric solution.
Our approach consequently estimates significantly more accurate trajectories, outperforming existing approaches by a large margin on challenging, uncalibrated datasets characterized by complex motions and severe visual aliasing.

While \ours\ represents a distinct leap forward for monocular reconstruction, its geometric precision does not yet match the strict tolerances of visual-inertial systems.
Mitigating long-range drift across increasingly longer sequences remains a formidable challenge, opening up exciting prospects for converting the world's abundant video streams into highly accurate 3D models for spatial computing and generative scene modeling.

\paragraph{Acknowledgments:} 
This work was supported by DSO National Laboratories with grant DSOCO24035.
We thank Xudong Jiang, Tobias Fischer, Shaohui Liu, and Haofei Xu for contributing to the early exploration of this topic.

\appendix
\begin{figure*}[t]
  \centering
  \includegraphics[width=\linewidth]{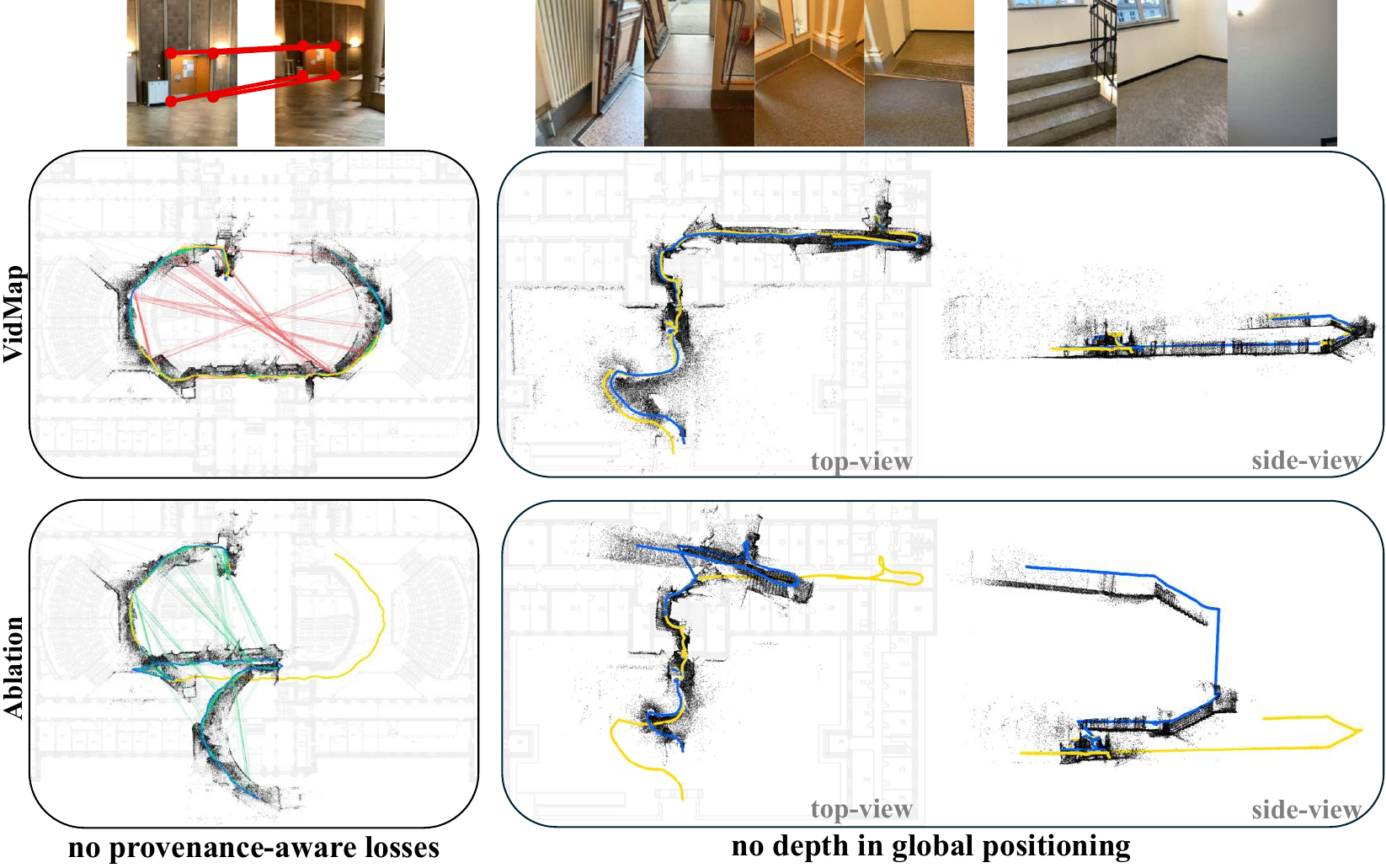}
  \caption{\textbf{Visual ablations.}
  Each column shows trajectory images above the corresponding reconstructions for \ours\ (top) and the ablation (bottom); GT trajectories are yellow and estimates are blue.
  Left: red loop-closure edges are rejected as outliers while green edges are kept as inliers.
  Without provenance-aware losses, \ours\ cannot downweight symmetries and the reconstruction collapses.
  Right: without depth priors in GP, challenging motion and degenerate geometry break global positioning and cause drift.}
  \label{fig:reconstruction_ablations}
\end{figure*}

\newpage
\section{Qualitative Results}
\label{sec:supp:qualitative}

\begin{figure}[!htbp]
    \centering
    \includegraphics[width=\linewidth]{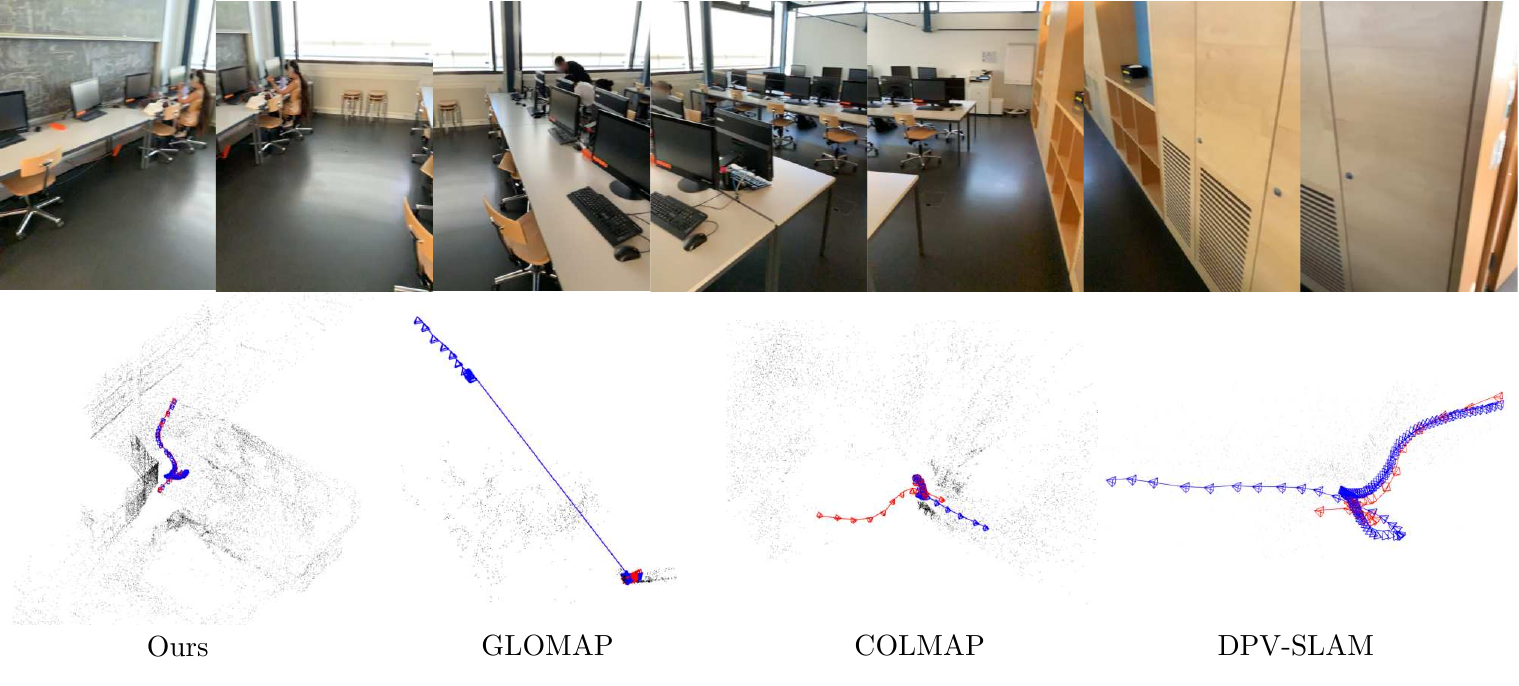}
    \caption{\textbf{Geometric priors resolve degenerate camera motion.} Pure rotations are frequent in unconstrained video and make triangulation impossible. Depth priors constrain the degenerate degrees of freedom: \ours\ recovers the room geometry while GLOMAP fragments, COLMAP registers few cameras, and DPV-SLAM drifts.
    \label{fig:supp:pure_rot}}
    \vspace{2em}
    \includegraphics[width=\linewidth]{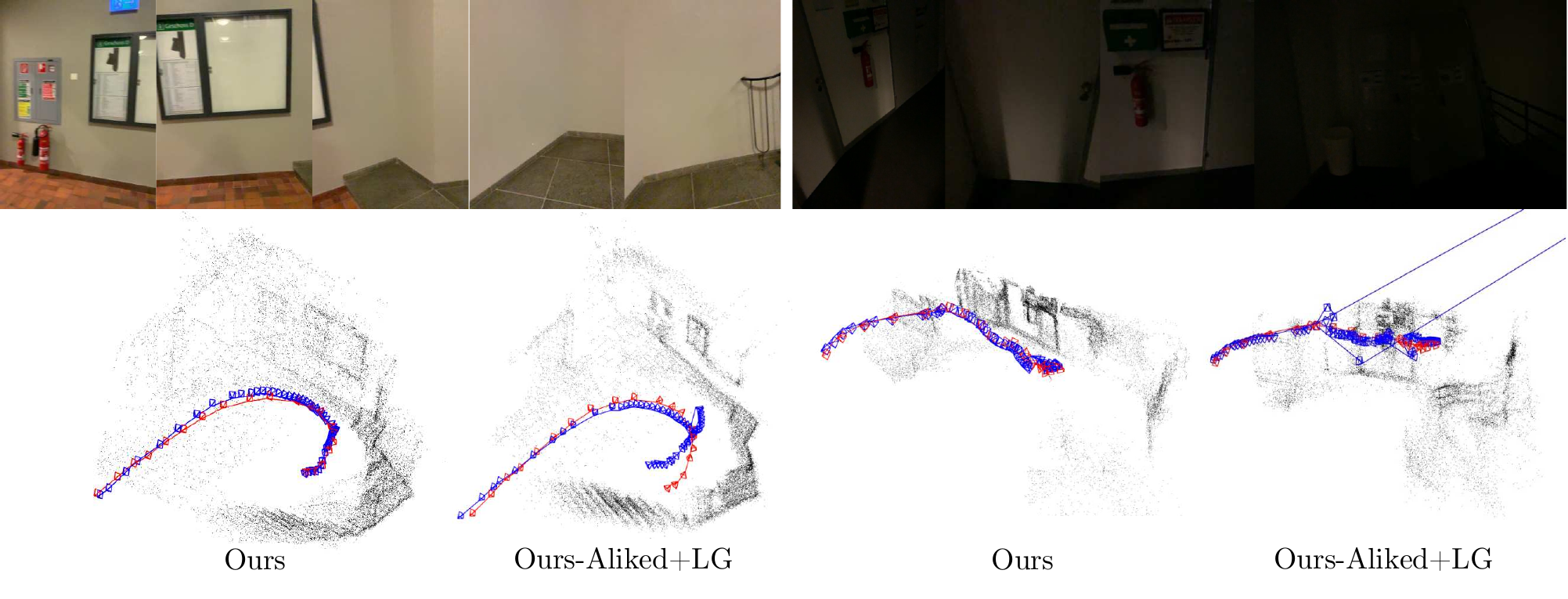}
    \caption{\textbf{Establishing tracks through textureless regions.} Textureless and feature-sparse scenes are abundant in unconstrained video. Left: the camera traverses a textureless wall; relying solely on salient keypoints (\ours-ALIKED+LG~\cite{zhao2023aliked}) leaves this segment under-constrained. Track propagation through dense matching maintains correspondences throughout. Right: a dark, feature-sparse room where \ours-ALIKED+LG cannot establish enough matches; thanks to loop closure the reconstruction recovers despite some lost frames.
    \label{fig:supp:trackloss}}
\end{figure}

\begin{figure}[!htbp]
    \centering
    \includegraphics[width=\linewidth]{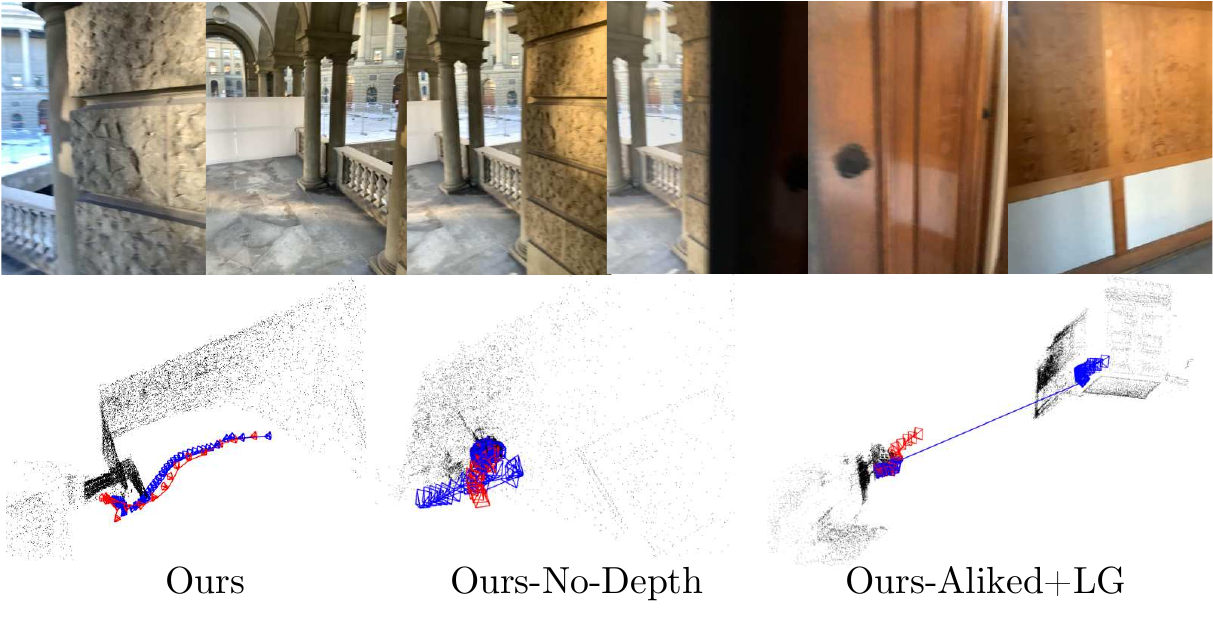}
    \vspace{-2.5em}
    \caption{\textbf{Difficult transition.}
    The camera moves through a doorway into a corridor, combining pure rotation with short tracks.
    Left: our approach recovers the trajectory.
    Middle: without geometric priors, the degenerate segment causes loss of tracking and collapse of the scale.
    Right: \ours-ALIKED+LG creates two disjoint reconstructions.
    Both components are needed: geometric priors alone cannot compensate for missing tracks and dense matching alone cannot resolve the degenerate geometry.
    \label{fig:supp:transition}}
    \vspace{1em}
    \includegraphics[width=\linewidth]{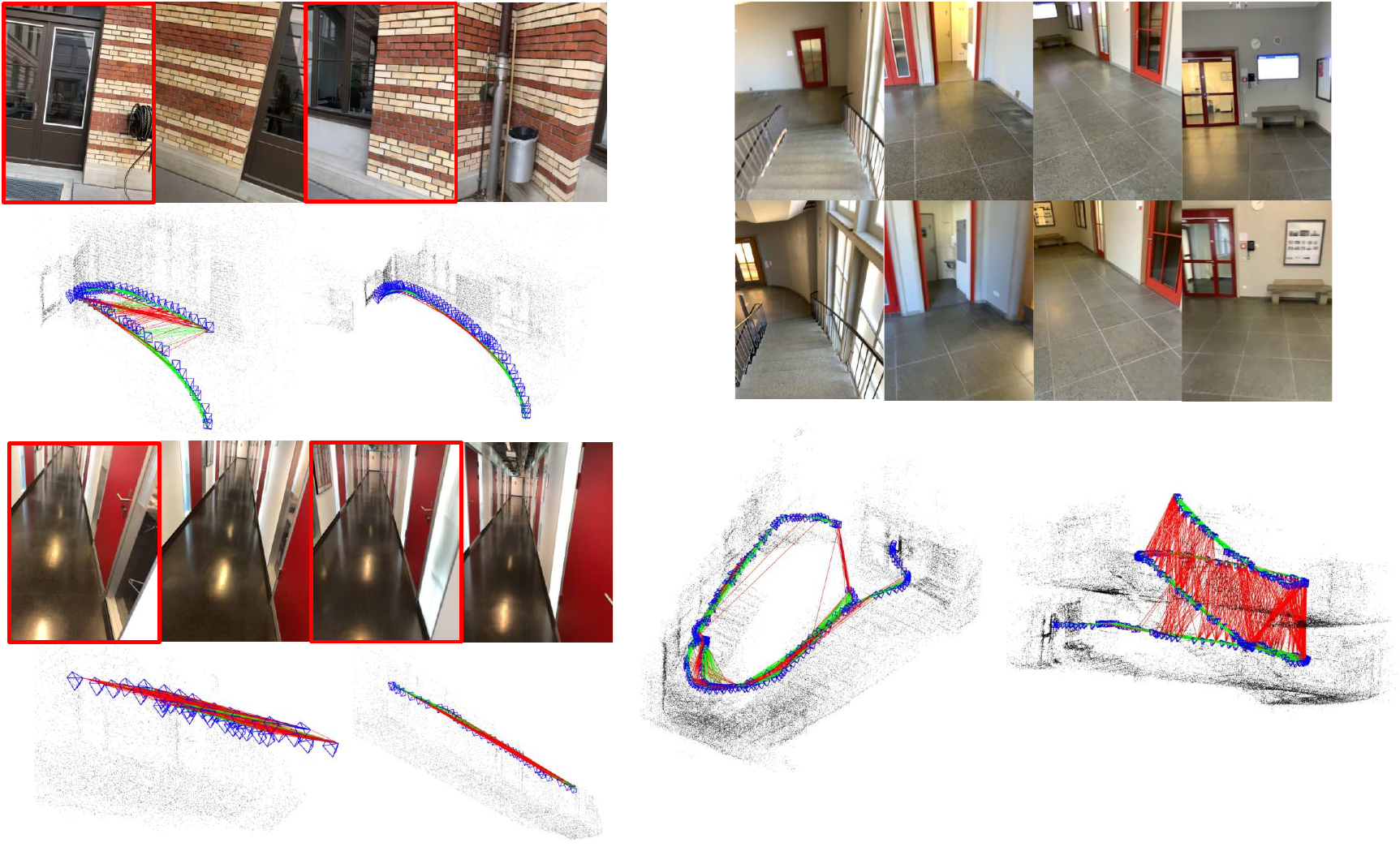}
    \vspace{-2.5em}
    \caption{\textbf{Challenges of local symmetries.}
    For three sequences, we show representative images that exhibit symmetries and visualize the camera poses and 3D point cloud estimated by standard SfM (left) and our approach (right).
    SfM treats all correspondences equally via transitivity and does not distinguish sequential and loop closure observations.
    Incorrect loop closures, visualized as red edges in the view graph, thus collapse the reconstruction (left).
    Our temporal track propagation enables provenance-dependent robust losses that suppress false closures (right).
    \label{fig:supp:local_lc}}
\end{figure}

\paragraph{Challenges in unconstrained video.}
Unconstrained video presents several recurring challenges for SfM. \textbf{Degenerate geometry}: pure rotations make triangulation impossible; depth priors resolve this (\cref{fig:supp:pure_rot}). \textbf{Textureless regions}: sparse matchers lack coverage; dense track propagation maintains correspondences through featureless segments (\cref{fig:supp:trackloss,fig:supp:transition}). \textbf{Local symmetries}: sequential matching confuses identical structures; temporal track ordering prevents jumps between instances (\cref{fig:supp:local_lc}). \textbf{Global symmetries}: at building scale, retrieval-based loop closure can propose false closures between similar-looking places; provenance-dependent robust losses downweight LC matches relative to sequential evidence. \textbf{Trajectory scale}: at longer trajectories these challenges compound; global optimization distributes error while causal methods accumulate drift. The key failure modes of \ours\ are discussed in \cref{sec:supp:limitations}.

\clearpage
\subsection{Limitations}
\label{sec:supp:limitations}

\begin{itemize}
    \item \textbf{Forward-motion keyframing.}
    During forward motion, many tracks remain near the epipole and exhibit little image displacement. The keyframing criterion can therefore select too few views when additional geometric support is needed, leaving global positioning underconstrained.

    \item \textbf{Interaction between weak geometry and visual aliasing.}
    Near-forward motion with few off-epipole tracks provides weak constraints on translation and depth. If visual aliasing yields many mutually consistent false loop closures, their aggregate influence can overwhelm the sequential constraints: Cauchy losses downweight each edge but do not eliminate its gradient. Global positioning may then converge to an incorrect solution.

    \item \textbf{Tracking failures.}
    Severe blur, occlusion, abrupt viewpoint changes, or low texture can terminate tracks. Because global optimization cannot recover missing correspondences, extensive track loss can cause drift or complete reconstruction failure.

    \item \textbf{Long-range drift.}
    Without reliable loop closures, small local errors can accumulate over very long sequences despite global optimization.

    \item \textbf{Dependence on learned priors.}
    Monocular depth and calibration priors can become unreliable for out-of-distribution imagery or cameras, limiting the accuracy of the reconstruction.
\end{itemize}

\begin{figure}[!htbp]
    \centering
    \includegraphics[width=\linewidth]{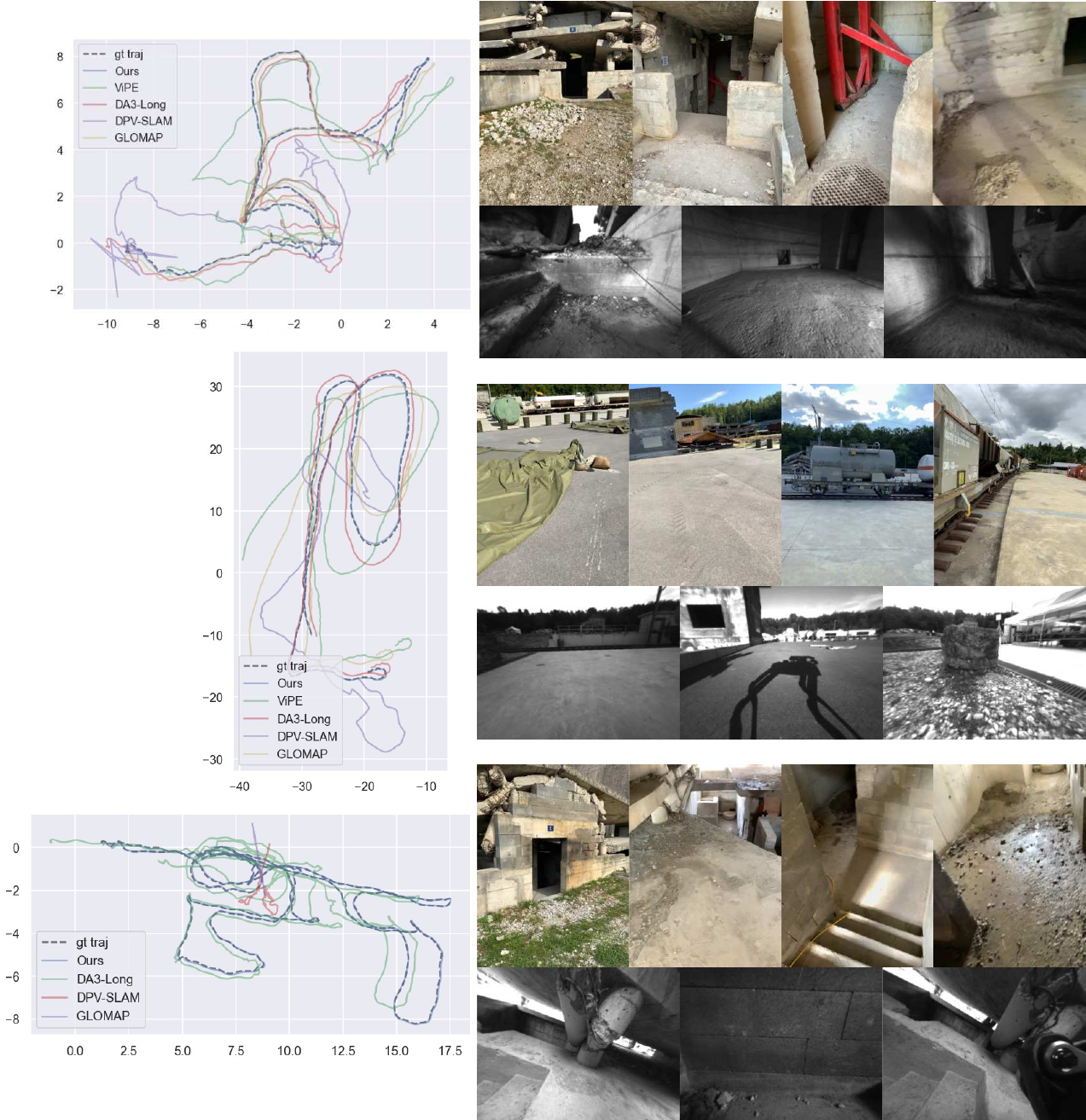}
    \caption{\textbf{Out-of-domain generalization on CroCoDL.} Three disaster-response scenes. Left: trajectories from iOS sequences for all baselines, aligned with GT (dashed). Right: corresponding scene imagery showing rubble, collapsed structures, and varying lighting. Top rows: iOS videos; bottom rows: quadruped capture. \ours\ remains precise while other methods struggle to generalize to these out-of-distribution environments.}
    \label{fig:supp:crocodl}
\end{figure}

\begin{figure}[!htbp]
    \centering
    \includegraphics[width=\linewidth]{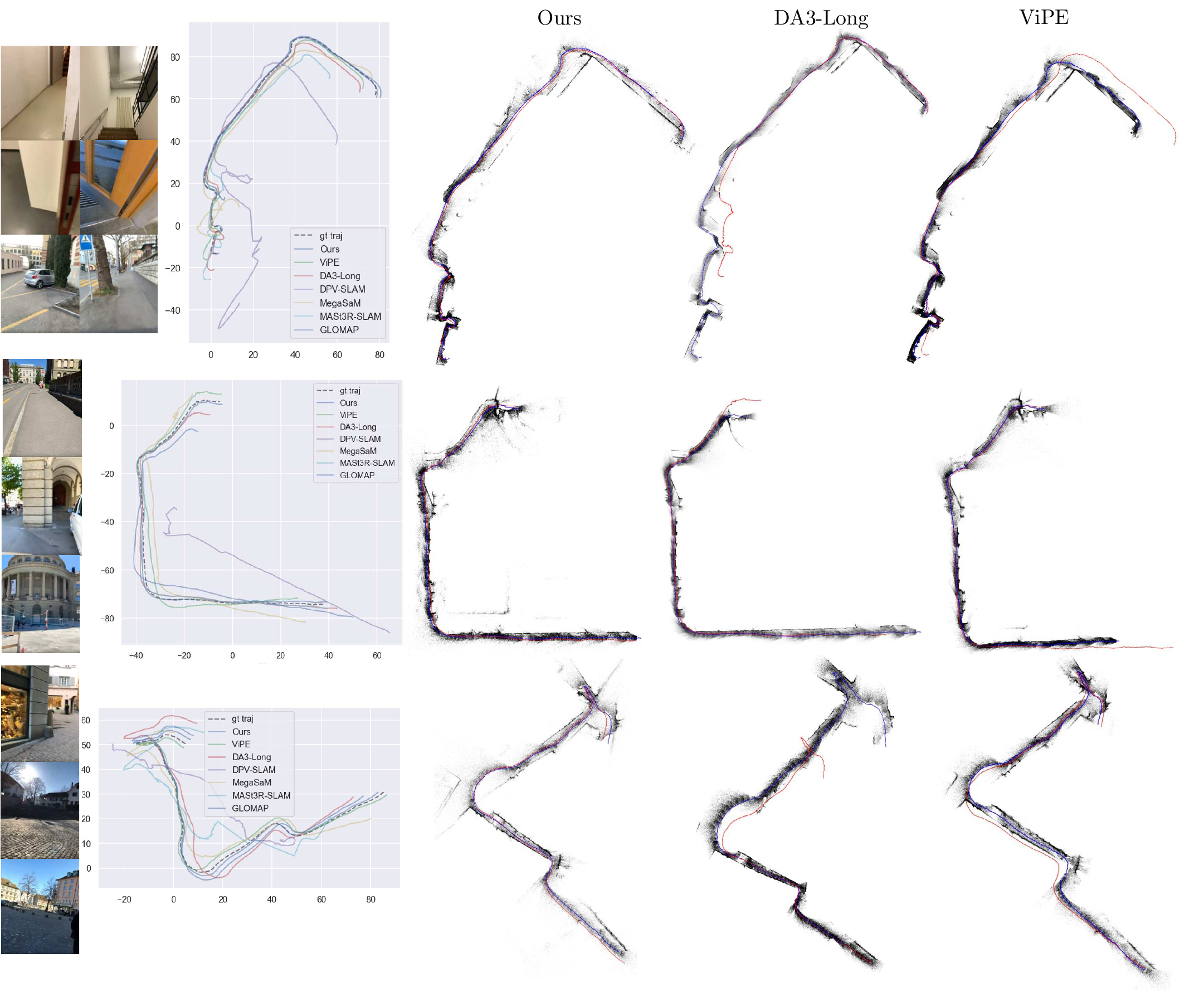}
    \caption{\textbf{Qualitative comparison on LaMAR.} Three scenes (CAB, HGE, LIN) with sample frames and bird's-eye trajectory plots (left) and 3D reconstructions with overlaid trajectories (right; red: GT, blue: estimated). \ours\ maintains consistent scale and topology; DA3-Long and ViPE exhibit progressive drift due to accumulated errors on longer sequences.}
    \label{fig:supp:lamar_comp}
\end{figure}

\begin{figure}[!htbp]
    \centering
    \includegraphics[width=\linewidth]{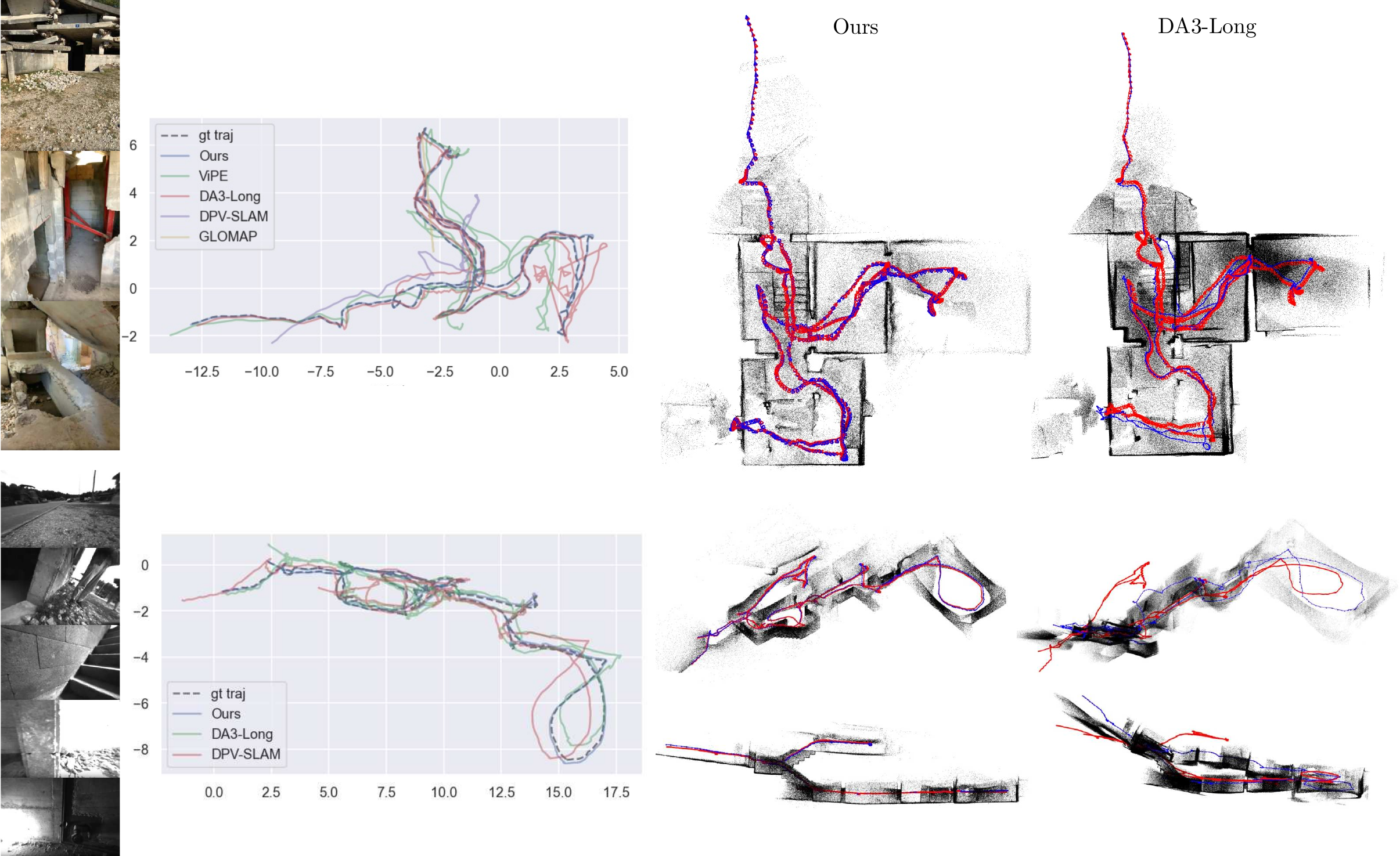}
    \caption{\textbf{Qualitative comparison on CroCoDL.} Two disaster-response scenes (iOS phone and quadruped robot). Bird's-eye trajectory plots (left) comparing all baselines; 3D reconstructions from top and side views (right; red: GT, blue: estimated). \ours\ generalizes to these out-of-distribution environments while baselines exhibit significant drift or fail to reconstruct.}
    \label{fig:supp:crocodl_compare}
\end{figure}

\paragraph{Visual comparisons.}
\Cref{fig:supp:lamar_comp,fig:supp:crocodl,fig:supp:crocodl_compare} compare \ours\ against existing approaches on LaMAR and CroCoDL. On LaMAR, causal methods (DA3-Long, ViPE) accumulate drift on longer trajectories while \ours\ maintains consistent scale and topology (\cref{fig:supp:lamar_comp}). On CroCoDL, \ours\ generalizes to out-of-distribution disaster-response environments where most baselines fail (\cref{fig:supp:crocodl,fig:supp:crocodl_compare}).

\section{Tracker Ablation}
\label{sec:supp:tracker}

\subsection{Track Quality (Epipolar Error)}
\label{sec:supp:tracker:quality}

Epipolar error across keyframe hops, evaluated with GT poses. Each tracker propagates ALIKED keypoints between keyframes using intermediate frames; tracks are then evaluated at keyframe positions against GT epipolar geometry. ETH3D SLAM provides sub-millimeter motion-capture GT, but LaMAR poses are derived from LiDAR-aligned SfM (${\sim}$cm-level accuracy). This GT noise inflates epipolar error equally for all methods, establishing a shared noise floor on LaMAR.

\begin{figure}[!htbp]
    \centering
    \includegraphics[width=\linewidth]{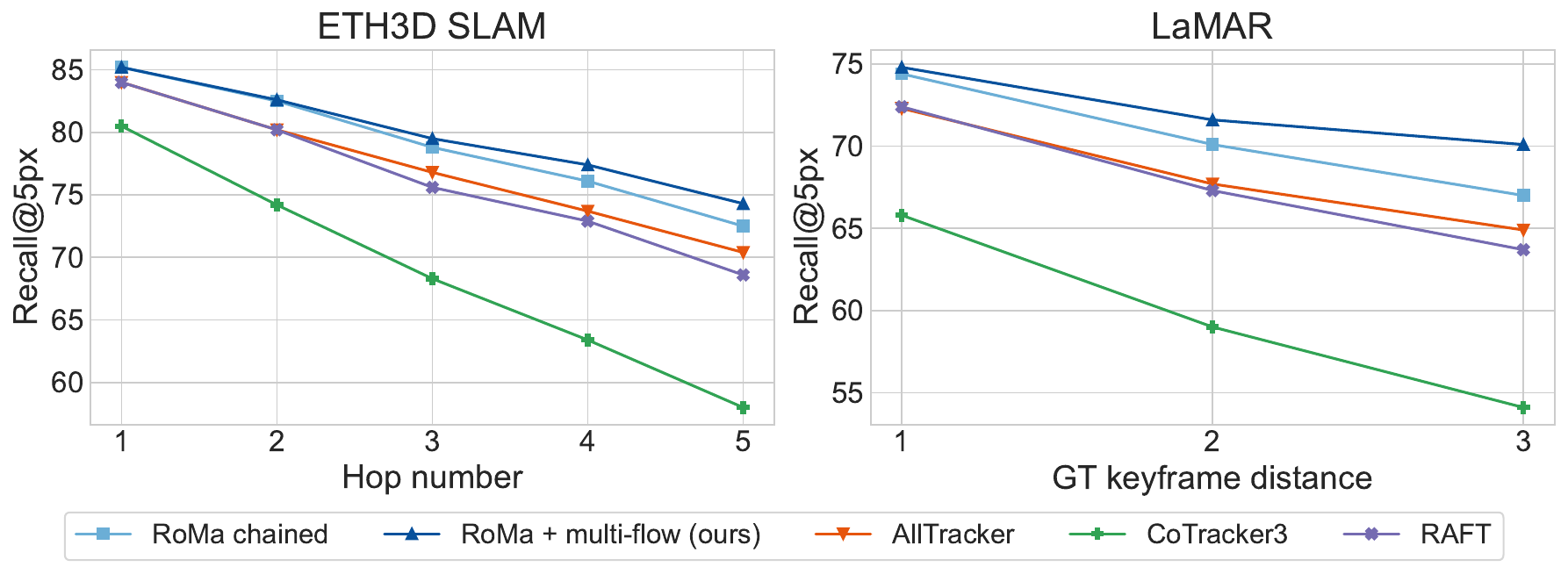}
    \caption{\textbf{Track quality across keyframe hops.} Recall@5px on ETH3D SLAM (left, by hop count) and LaMAR (right, by GT keyframe distance). RoMa already outperforms all flow-based trackers when used as a simple chained matcher; multi-flow propagation further widens the gap, confirming that the correction step contributes beyond the choice of backbone.}
    \label{fig:supp:recall_combined}
\end{figure}

\begin{figure}[!htbp]
    \centering
    \includegraphics[width=\linewidth]{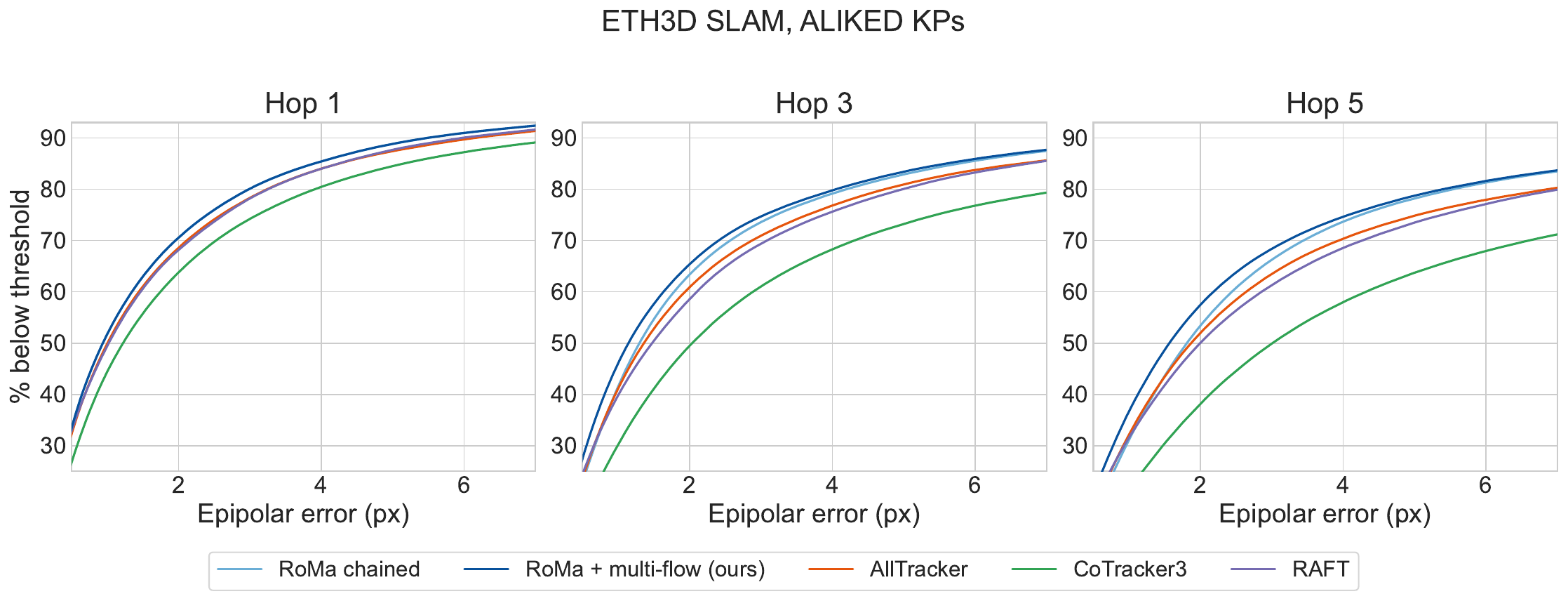}\\[4pt]
    \includegraphics[width=\linewidth]{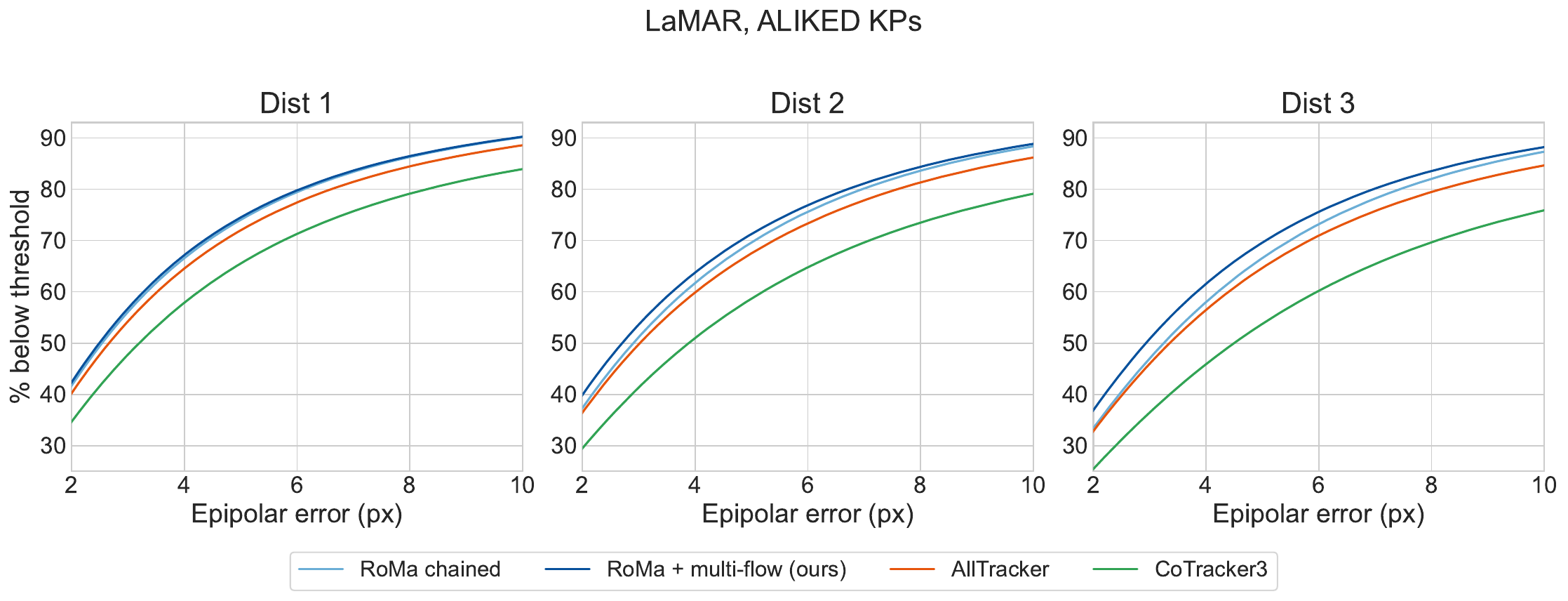}
    \caption{\textbf{Epipolar error CDF per dataset.} Cumulative distribution of epipolar error for ALIKED keypoints. Top: ETH3D SLAM, by keyframe hop count. Bottom: LaMAR, by GT keyframe distance bin (Dist 1--3, increasing displacement). All methods perform comparably at short range; as hops or distance increase, flow-based trackers shift rightward as drift compounds, while RoMa with multi-flow maintains the tightest distribution. The higher noise floor on LaMAR reflects its cm-level GT accuracy.}
    \label{fig:supp:cdf_eth3d}
    \label{fig:supp:cdf_lamar}
\end{figure}

\paragraph{Discussion.}
RoMa consistently outperforms all flow-based trackers~\cite{neoral2024mft} on both datasets across all hop distances. Multi-flow propagation further widens this gap as drift compounds at longer hops. CoTracker3, trained on synthetic data to model dynamic objects, fails to accurately track rigid 3D structure; modern deep trackers like AllTracker~\cite{harley2025alltracker} improve over two-view RAFT but a significant accuracy gap to global matching remains. These averages understate the problem: in unconstrained video, the camera can undergo rapid viewpoint changes (\eg head rotation with a head-mounted device), producing large scene-space displacements between consecutive keyframes. Flow-based methods propagate matches locally and cannot recover from such displacements, whereas RoMa establishes correspondences from scratch per pair. The end-to-end comparison (\cref{sec:supp:tracker:e2e}) confirms this: replacing RoMa with a tracker degrades reconstruction quality, particularly on sequences with abrupt viewpoint changes. \ours-ALIKED+LG suffers from a complementary failure: sparse keypoint detectors lack repeatability on textureless regions, producing few or no matches where dense propagation maintains coverage. Our full pipeline combines the robustness of global matching with the coverage of dense tracking.

\subsection{Tracker Backbone Comparison}
\label{sec:supp:tracker:e2e}

\paragraph{Tracker variants.}
For RAFT, AllTracker, and CoTracker3, we replace only the track propagation backbone: each tracker propagates ALIKED keypoints between keyframes via intermediate frames using sliding-window extraction, while keyframe selection, loop closure matching, depth extraction, and global mapping remain identical to \ours. \ours~(equivalent) uses RoMa as a chained tracker (frame-to-frame propagation without multi-flow correction), sampling only ALIKED keypoints to isolate the backbone as the sole variable.

\paragraph{\ours-ALIKED+LG.}
This variant replaces dense tracking with sparse feature matching throughout. ALIKED features are extracted per keyframe and matched with LightGlue on consecutive pairs; tracks are chained through transitivity, accumulating drift analogously to flow-based propagation. Window-based matching provides geometric verification and loop closure correspondences.

\begin{figure}[!htbp]
    \centering
    \includegraphics[width=\linewidth]{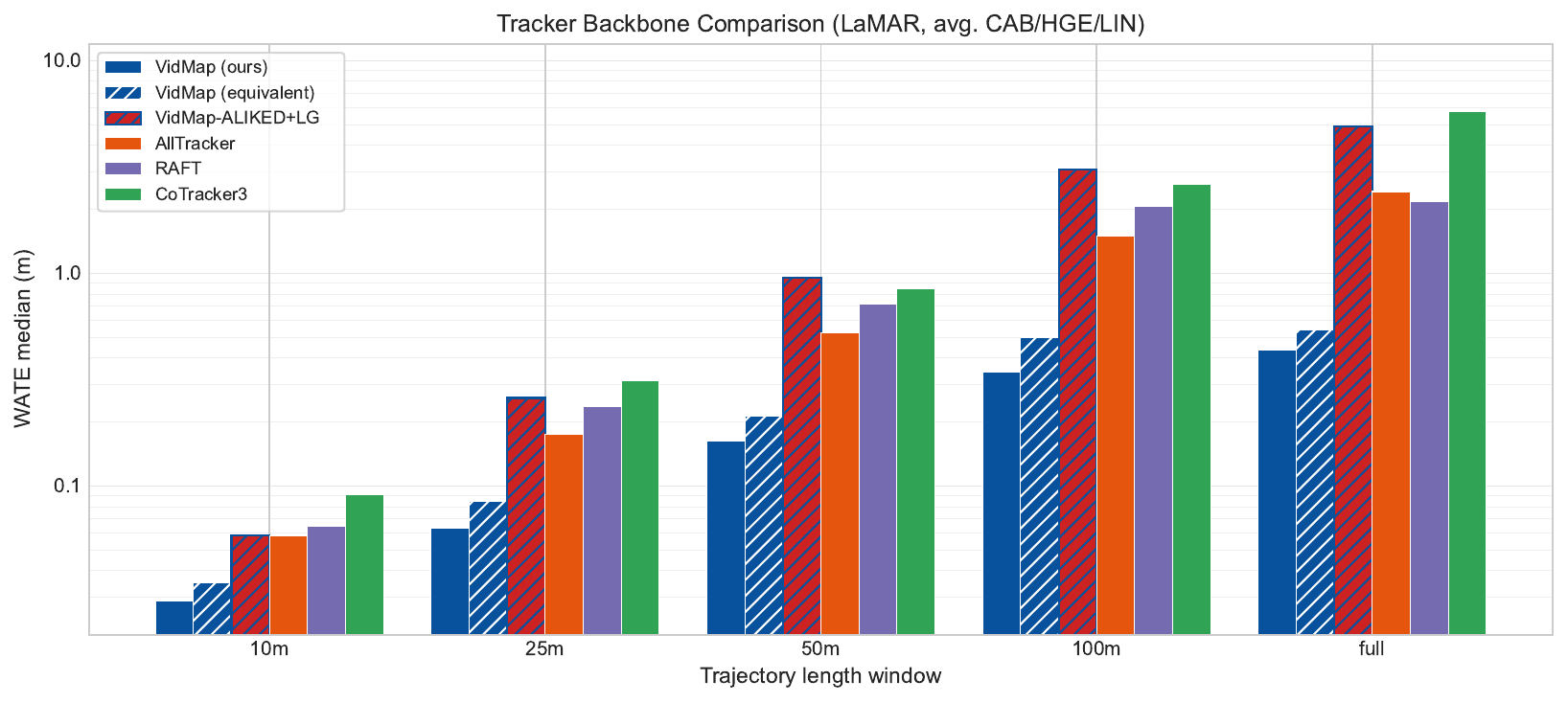}
    \caption{\textbf{Tracker backbone comparison on LaMAR.} Median WTE (m) averaged across scenes at increasing trajectory-length windows. Replacing RoMa with flow-based trackers degrades reconstruction quality at all scales. \ours-ALIKED+LG, relying on keypoint repeatability, loses coverage on textureless regions where dense propagation maintains correspondences.}
    \label{fig:supp:e2e}
\end{figure}

\section{Dataset Details}
\label{sec:supp:datasets}

\subsection{Dataset Setup}
\label{sec:supp:datasets:setup}

We evaluate four benchmarks spanning diverse capture devices, environments, and motion patterns. \Cref{tab:supp:datasets} summarizes key properties.

\begin{table}[!htbp]
\centering
\caption{\textbf{Overview of evaluation datasets.} Four benchmarks spanning complex camera motion over long indoor-outdoor trajectories (LaMAR), out-of-distribution disaster-response environments with iOS and quadruped capture (CroCoDL), precise small-scale motion (ETH3D), and MAV flight in grayscale (EuRoC).}
\label{tab:supp:datasets}
\setlength{\tabcolsep}{2pt}
\tiny
\begin{tabular}{l cccc}
\toprule
& \textbf{LaMAR} & \textbf{ETH3D-SLAM} & \textbf{EuRoC} & \textbf{CroCoDL} \\
\midrule
Scenes / sites & 3 (CAB, HGE, LIN) & 55 sequences & 11 sequences & 4 sites \\
Devices & iOS (iPhone/iPad) & Global-shutter RGB & Grayscale stereo MAV & iOS + quadruped \\
Resolution & $1920{\times}1440$ & ${\sim}740{\times}460$ & $752{\times}480$ & $1440{\times}1920$ / $640{\times}480$ \\
Color & RGB & RGB & Grayscale & RGB / Grayscale \\
Ground truth & LiDAR-aligned & Motion capture & Vicon + Leica & LiDAR-aligned \\
Sequences eval'd & 50 & 55 & 11 & 88 \\
\bottomrule
\end{tabular}
\end{table}

\paragraph{LaMAR~\cite{sarlin2022lamar}.}
Three scenes in Zurich (CAB, HGE: indoor + outdoor; LIN: outdoor) captured with iOS devices, totaling 50 sequences with GT poses (12 CAB, 17 HGE, 21 LIN). GT is derived from NavVis LiDAR scans and SfM bundle adjustment at ${\sim}$cm-level accuracy, but exists only for a sparse subset of keyframes per sequence that do not span the full video. LaMAR GT keyframes are spaced at larger displacement than typical SLAM keyframes, making the two not directly comparable. For balanced evaluation, we sample trajectories spanning 300 GT keyframes each.

\paragraph{ETH3D-SLAM~\cite{schops2019badslam}.}
55 of 61 sequences evaluated; 6 dark sequences excluded due to near-black imagery.

\paragraph{EuRoC~\cite{burri2016euroc}.}
11 MAV sequences in grayscale at $752{\times}480$, evaluated monocularly (left camera only). Grayscale imagery is challenging for learned depth models trained on color.

\paragraph{CroCoDL~\cite{Blum_2025_CVPR}.}
Four disaster-response sites with 62 iOS and 14 quadruped sessions. GT from the LaMAR pipeline~\cite{sarlin2022lamar} at ${\sim}$10\,cm accuracy. Quadruped sequences use the left-facing camera from rosbag; sessions with significant buffering artifacts are excluded.

\section{Evaluation Details}
\label{sec:supp:eval}

\subsection{Evaluation Protocol}
\label{sec:supp:eval:protocol}

\paragraph{AUC.} On ETH3D-SLAM and EuRoC, we report translation-error area under the curve (AUC) at thresholds $\{0.05, 0.1, 0.5, 1.0, 10\}$\,m (higher is better). Poses are globally aligned to GT with COLMAP's robust Sim(3) aligner. AUC at threshold $\tau$:
\begin{equation}
    \text{AUC}@\tau = \frac{1}{\tau} \int_0^\tau \text{recall}(t)\, dt
    \label{eq:auc}
\end{equation}
where $\text{recall}(t) = |\{i : \text{TE}_i \leq t\}| / N$ is the fraction of poses with translational error $\leq t$. 

\paragraph{W-AUC.} On LaMAR and CroCoDL, we first compute per-frame windowed translation error (WTE): for each frame, a $W$-meter window is centered on it, local Sim(3) alignment is computed within the window, and the center-frame translation error is stored. W-AUC then applies the same AUC computation to these per-frame translation errors for $W \in \{10, 25, 50, 100\}$\,m and the full trajectory, using an error threshold equal to 5\% of each window length.

\paragraph{Why W-AUC over AUC.}
Global Sim(3) alignment is dominated by long-range drift on long trajectories and can obscure local reconstruction quality. W-AUC evaluates translation errors after local windowed alignment, making it better suited to long videos.

\subsection{Baseline Details}
\label{sec:supp:eval:baselines}

We use the official implementations unless otherwise noted.
Since SLAM approaches typically assume fixed intrinsics per video, in the calibrated setting we provide them with per-frame ground truth intrinsics, which generally improves their results.

\begingroup
\emergencystretch=.85em

\textbf{COLMAP / GLOMAP.} We evaluate both systems with two matching variants: SuperPoint+LightGlue (COLMAP/GLOMAP-LG) and SuperPoint+RoMA v2 (COLMAP/GLOMAP-RoMA), all with sequential matching in a local window and across loop-closures retrieved with MegaLoc~\cite{megaloc}.
We use the tuned COLMAP configuration from MP-SfM~\cite{pataki2025mpsfm}, which is more permissive toward low triangulation angles, letting it reconstruct a larger proportion of trajectories in the LaMAR dataset.

\textbf{MP-SfM.} We use the same matching strategy as COLMAP and GLOMAP, described above.

\textbf{MegaSaM.} In the calibrated setting, we leverage the per-frame ground truth intrinsics in the SLAM backend and when estimating monocular depth maps with UniDepth.

\textbf{MASt3R-SLAM.} We augment MASt3R-SLAM with two modifications that significantly improve its results:
(i) MASt3R-SLAM's keyframe selection threshold can reject frames for too long on complex sequences, eventually resulting in tracking loss from which the system never recovers. Instead, we retroactively register the last frame before tracking failure as a keyframe.
(ii) Because not every keyframe on LaMAR has a ground truth pose, we additionally register frames at exactly those timestamps that do to ensure fair evaluation coverage.

\par
\endgroup
\section{Implementation Details}
\label{sec:supp:impl}

\subsection{Derivations}
\label{sec:supp:impl:derivations}

\subsubsection{Observation Covariance in Bearing Space.}
The isotropic pixel-noise covariance $\sigma^2 I_2$ is propagated through unprojection and subsequent normalization to a unit bearing, yielding an anisotropic covariance in bearing space. The pinhole unprojection Jacobian $D$ and the resulting image-space covariance are
\begin{equation}
    D = \operatorname{diag}(1/f_x,\, 1/f_y), \quad
    \Sigma_{\text{img}} = \sigma^2 \operatorname{diag}\!\left(\frac{1}{f_x^2},\, \frac{1}{f_y^2}\right).
\end{equation}
Normalizing $\mathbf{p}$ to the unit bearing $\mathbf{b} = \mathbf{p}/\|\mathbf{p}\|$ yields the Jacobian
\begin{equation}
    J = b_z\!\left(I_3 - \mathbf{b}\,\mathbf{b}^\top\right),
\end{equation}
where $b_z$ denotes the $z$-component of $\mathbf{b}$.
Let $J_2 \in \mathbb{R}^{3 \times 2}$ contain the first two columns of $J$. The resulting bearing covariance is
\begin{equation}
    \Sigma^v_{ik} = J_2\,\Sigma_{\text{img}}\,J_2^\top \in \mathbb{R}^{3 \times 3}.
\end{equation}
This covariance has rank 2, with null direction $\mathbf{b}$; it is direction-dependent and scales with $b_z^2$. For GP, we use a diagonal approximation to this covariance, and set the third standard deviation to the mean of the first two, $\sigma_z = (\sigma_x + \sigma_y)/2$.

\subsubsection{Density-Weighted Keypoint Sampling.}
We adapt the density-balanced sampling used in RoMaV2~\cite{edstedt2025romav2} to the combined set of propagated and candidate keypoints, computing a Gaussian kernel density estimate $\delta(\mathbf{x})$ with bandwidth $h$:
\begin{equation}
    \delta(\mathbf{x}) = \sum_{j} \phi_h(\mathbf{x}, \mathbf{p}_j) + \sum_{l} \phi_h(\mathbf{x}, \mathbf{q}_l), \quad
    \phi_h(\mathbf{x}, \mathbf{y}) = \exp\!\left(-\frac{\|\mathbf{x} - \mathbf{y}\|^2}{2h^2}\right).
\end{equation}
The corresponding acceptance probability is
\begin{equation}
    p(\mathbf{x}) \propto \frac{1}{\delta(\mathbf{x}) + 1}.
\end{equation}

\subsubsection{Track Density Thinning.}
\label{sec:supp:impl:track-thinning}
After epipolar filtering, we recompute $\delta(\mathbf{x})$ over the surviving observations and retain each with probability
\begin{equation}
    p_{\text{survive}}(\mathbf{x}) = \exp\!\left(-\beta \cdot \left(\delta(\mathbf{x}) - 1\right)\right)
    \enspace,
\end{equation}
where $\beta>0$ controls the thinning strength and subtracting one removes the observation's self-contribution. This preferentially removes observations in dense regions.

\subsection{Keyframing}
\label{sec:supp:impl:keyframing}

A new keyframe is inserted when the fraction of pixels with certainty above $\tau_c$ and accumulated displacement below $\tau_0$ drops below the target coverage $\tau_f$.
ALIKED keypoints are propagated over a sliding window of $W$ keyframes, up to $N$ per image.
Matching resolutions and window parameters are listed in \cref{tab:supp:keyframing}.

\begin{table}[!htbp]
\centering
\caption{\textbf{Keyframing and track extraction parameters.}}
\label{tab:supp:keyframing}
\scriptsize
\begin{tabular}{lccc}
\toprule
Parameter & Symbol & Value & Role \\
\midrule
Certainty threshold & $\tau_c$ & 0.01 & Min RoMa match certainty to count as tracked \\
Displacement threshold & $\tau_0$ & 0.12 & Min motion to trigger new keyframe \\
Target coverage & $\tau_f$ & 0.4 & KF inserted when tracked fraction drops below this \\
Max keypoints/image & $N$ & 1200 & Track budget per keyframe \\
Low-res RoMa input & -- & $560{\times}560$ & Coarse matching and keyframing \\
Sequential RoMa input & -- & 1200 & Longest side for track propagation \\
LC RoMa input & -- & 800 & Longest side for loop-closure matching \\
Sliding window & $W$ & 9 & Keyframes over which tracks are propagated \\
Window stride & -- & 2 & Match every second preceding keyframe \\
\bottomrule
\end{tabular}
\end{table}

\subsection{Depth-Regularized Relative Pose (MDRP)}
\label{sec:supp:impl:mdrp}

Per-KF-pair relative pose via PoseLib \texttt{estimate\_monodepth\_pose}. Jointly solves rotation, translation, and metric scale from correspondences + monocular depth.

MDRP handles degenerate configurations (planar scenes, pure rotation, forward motion) where the 5-point algorithm fails. Within a pipeline already designed for robustness, the marginal improvement on aggregate metrics is small.

\begin{table}[!htbp]
\centering
\caption{\textbf{MDRP ablation on LaMAR-HGE.} Median WTE (m, $\downarrow$) with and without depth-regularized relative pose estimation. MDRP handles degenerate geometry robustly, though improvements within the full pipeline are marginal.}
\label{tab:supp:mdrp}
\scriptsize
\begin{tabular}{l ccccc}
\toprule
& 10\,m & 25\,m & 50\,m & 100\,m & full \\
\midrule
Ours             & \textbf{0.03} & 0.06 & \textbf{0.14} & \textbf{0.38} & \textbf{0.43} \\
Ours (no MDRP) & 0.03 & 0.06 & 0.17 & 0.39 & 0.45 \\
\bottomrule
\end{tabular}
\end{table}

\subsection{Global Positioning Loss Schedule}
\label{sec:supp:impl:gp}

Global positioning solves all cameras jointly from random initialization, where poses are arbitrary and 3D points may lie behind cameras. Applying robust losses too early suppresses the large residuals from this random init: degenerate parts of the sequence have no gradient from matches since all observations are initially outliers, and geometric priors are similarly zeroed out. In the early GP pass, pairwise depth consistency checking first flags depth/match inconsistencies as outliers, preventing the subsequent L2 depth loss from introducing harmful gradients. L2 then ensures all remaining observations contribute gradient. Once geometry converges, a late GP pass replaces L2 with robust losses, suppressing true outliers while trusting clean matches more tightly.

\subsection{Bundle Adjustment with Graduated Convexity}
\label{sec:supp:impl:ba}

GP yields less precise absolute pose estimates than incremental SfM, so early BA follows GLOMAP with moderate reprojection filtering and loose robust losses; tight thresholds would discard valid observations at this stage. Once geometry stabilizes, a second BA pass tightens reprojection error filtering and applies graduated convexity: robust loss thresholds are progressively tightened, trusting inlier matches increasingly as the reconstruction matures. Skipping this refinement degrades performance at strict thresholds, where sub-cm accuracy requires precise point filtering and tight robust losses.

\begin{table}[t]
\centering
\caption{\textbf{Graduated BA refinement ablation.} AUC ($\uparrow$) at various thresholds. The second refinement pass improves strict-threshold accuracy by $+8.2$ on ETH3D (5\,cm) and $+10.0$ on EuRoC; on LaMAR gains appear at coarser windows where increased precision reduces drift.}
\label{tab:supp:ba_ablation}
\setlength{\tabcolsep}{2pt}
\scriptsize
\begin{tabular}{l ccccc ccccc ccccc}
\toprule
& \multicolumn{5}{c}{LaMAR (sample-300)} & \multicolumn{5}{c}{ETH3D} & \multicolumn{5}{c}{EuRoC} \\
\cmidrule(lr){2-6} \cmidrule(lr){7-11} \cmidrule(lr){12-16}
& .05 & .1 & .5 & 1 & 10 & .05 & .1 & .5 & 1 & 10 & .05 & .1 & .5 & 1 & 10 \\
\midrule
Ours                     & 3.7 & 7.9 & 35.3 & 57.0 & 93.3 & 67.5 & 76.8 & 89.7 & 93.1 & 99.1 & 22.2 & 53.8 & 90.7 & 95.4 & 99.6 \\
Ours (no graduated BA)   & 3.5 & 7.0 & 29.2 & 50.8 & 91.8 & 59.3 & 71.5 & 89.5 & 93.7 & 98.9 & 12.2 & 41.4 & 87.5 & 93.8 & 99.4 \\
\bottomrule
\end{tabular}
\end{table}

\bibliographystyle{splncs04}
\bibliography{main}

\end{document}